\documentclass[sigconf]{acmart}

\usepackage{booktabs} %
\usepackage{enumitem}
\usepackage{bbm}
\usepackage{amsmath}
\usepackage{placeins}
\usepackage[ruled,vlined, algosection, procnumbered]{algorithm2e}
\usepackage{graphicx}
\usepackage{subfig}
\usepackage{afterpage}
\usepackage{ragged2e}
\usepackage{hyperref}

\usepackage{chngcntr}
\counterwithin{figure}{section}

\usepackage{caption}
\DeclareCaptionLabelFormat{cont}{#1~#2\alph{ContinuedFloat}}
\setcopyright{rightsretained}
\acmConference[KDD'18]{ACM conference}{August 2018}{London, United Kingdom} 
\acmYear{2018}
\copyrightyear{2018}

\acmArticle{4}

\begin{document}
\title{Extremely Fast Decision Tree}

\author{Chaitanya Manapragada}
\orcid{1234-5678-9012}
\affiliation{%
  \institution{Monash University}
  \streetaddress{Clayton}
  \city{Victoria} 
  \state{Australia} 
  \postcode{3800}
}
\email{chait.m@monash.edu}

\author{Geoffrey I. Webb}
\orcid{1234-5678-9012}
\affiliation{%
	\institution{Monash University}
	\streetaddress{Clayton}
	\city{Victoria} 
	\state{Australia} 
	\postcode{43017-6221}
}
\email{geoff.webb@monash.edu}

\author{Mahsa Salehi}
\orcid{1234-5678-9012}
\affiliation{%
	\institution{Monash University}
	\streetaddress{Clayton}
	\city{Victoria} 
	\state{Australia} 
	\postcode{43017-6221}
}
\email{mahsa.salehi@monash.edu}

\renewcommand{\shortauthors}{C. Manapragada et al.}

\begin{abstract}
We introduce a novel incremental decision tree learning algorithm, Hoeffding Anytime Tree, that is statistically more efficient than the current state-of-the-art, Hoeffding Tree. We demonstrate that an implementation of Hoeffding Anytime Tree---``Extremely Fast Decision Tree'', a minor modification to the MOA implementation of Hoeffding Tree---obtains significantly superior prequential accuracy on most of the largest classification datasets from the UCI repository. Hoeffding Anytime Tree produces the asymptotic batch tree in the limit, is naturally resilient to concept drift, and can be used as a higher accuracy replacement for Hoeffding Tree in most scenarios, at a small additional computational cost.

\end{abstract}



\maketitle

\section{Introduction}


We present a novel stream learning algorithm, Hoeffding Anytime Tree (HATT)\footnote{In order to distinguish it from Hoeffding Adaptive Tree, or HAT \cite{bifet2009adaptive}}. The de facto standard for learning decision trees from streaming data is Hoeffding Tree (HT) \cite{domingos2000mining}, which is used as a base for many state-of-the-art drift learners \cite{hulten2001mining, bifet2009adaptive, brzezinski2014reacting, Santos2014, Barros2016, bifet2009new, hoeglinger2007use}. We improve upon HT by learning more rapidly and guaranteeing convergence to the asymptotic batch decision tree on a stationary distribution.

Our implementation of the Hoeffding Anytime Tree algorithm, the Extremely Fast Decision Tree (EFDT), achieves higher prequential accuracy than the Hoeffding Tree implementation Very Fast Decision Tree (VFDT) on many standard benchmark tasks.

HT constructs a tree incrementally, delaying the selection of a split at a node until it is confident it has identified the best split, and never revisiting that decision. In contrast, HATT seeks to select and deploy a split as soon as it is confident the split is useful, and then revisits that decision, replacing the split if it subsequently becomes evident that a better split is available.

The HT strategy is more efficient computationally, but HATT is more efficient statistically, learning more rapidly from a stationary distribution and eventually learning the asymptotic batch tree if the distribution from which the data are drawn is stationary. Further, false acceptances are inevitable, and since HT never revisits decisions, increasingly greater divergence from the asymptotic batch learner results as the tree size increases (Sec. \ref{sec:relatedwork}).

\begin{figure}[t]
	
	\subfloat[a][VFDT: the current de facto standard for incremental tree learning
	]{
		\includegraphics[height=1.4in, width=3.6in]{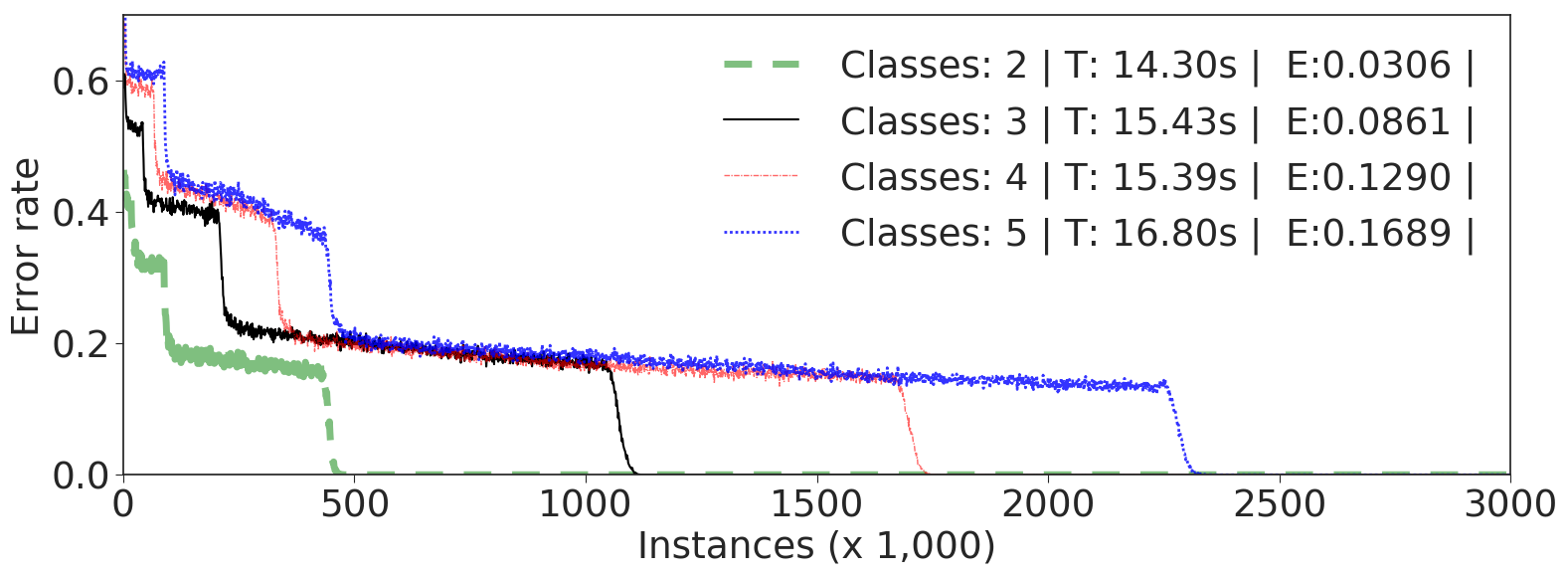}}

	\vspace*{-5pt}\subfloat[b][	EFDT: our more statistically efficient variant]{
		
		\includegraphics[height=1.4in, width=3.6in]{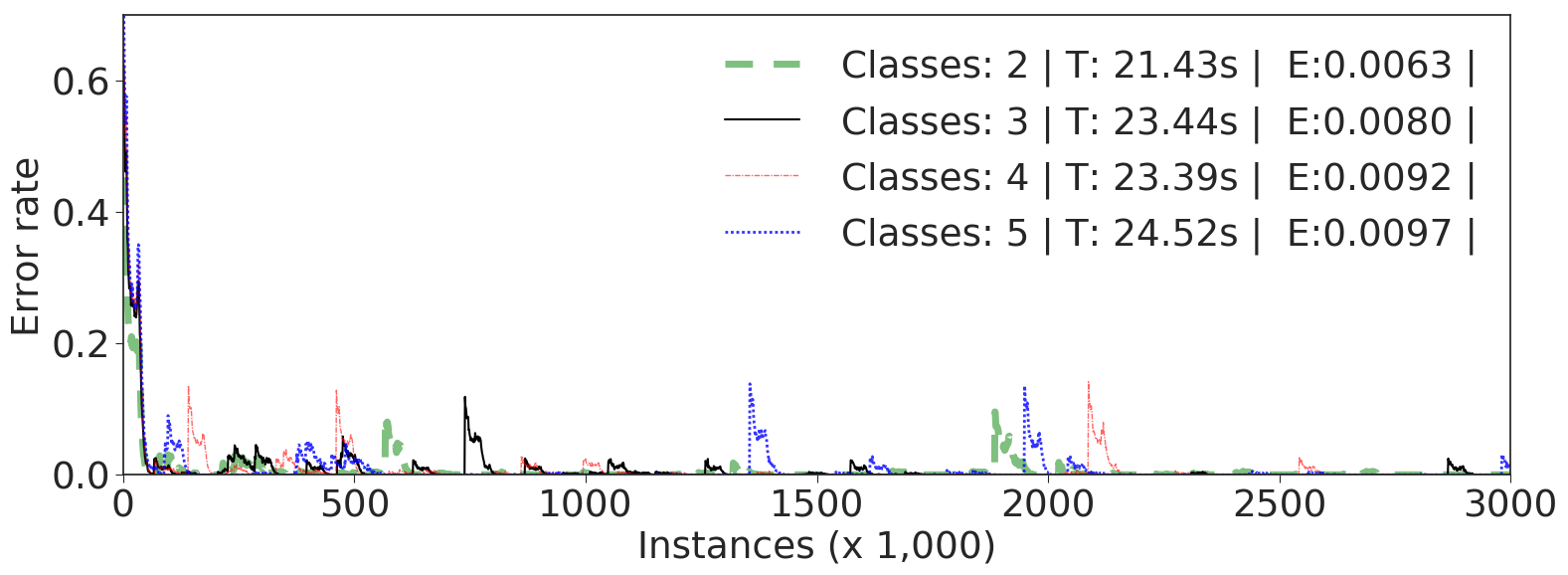} }

	\caption{The evolution of prequential error over the duration of a data stream. For each learner we plot error for 4 different levels of complexity, resulting from varying the number of classes from 2 to 5. The legend includes time in CPU seconds (T) and the total error rate over the entire duration of the stream (E). This illustrates how EFDT learns much more rapidly than VFDT and is less affected by the complexity of the learning task, albeit incurring a modest computational overhead to do so. The data are generated by
		MOA RandomTreeGenerator, 5 classes, 5 nominal attributes, 5 values per attribute, 10 stream average.}
	\label{fig:intro}
\end{figure}


In Fig. \ref{fig:intro}, we observe VFDT taking longer and longer to learn progressively more difficult concepts obtained by increasing the number of classes. EFDT learns all of the concepts very quickly, and keeps adjusting for potential overfitting as fresh examples are observed. 

In Section \ref{sec:performance}, we will see that EFDT continues to retain its advantage even 100 million examples in, and that EFDT achieves significantly lower prequential error relative to VFDT on the majority of benchmark datasets we have tested. VFDT only slightly outperforms EFDT on three synthetic physics simulation datasets---Higgs, SUSY, and Hepmass. 

\section{Background}\label{sec:bg}

Domingos and Hulten presented one of the first algorithms for incrementally constructing a decision tree in their widely acclaimed work, ``Mining High-Speed Data Streams'' \cite{domingos2000mining}.

Their algorithm is the Hoeffding Tree (Table \ref{table:vfdt}), which uses the \textit{Hoeffding Bound}. For any given potential split, Hoeffding Tree checks whether the difference of averaged information gains of the top two attributes is likely to have a positive mean---if so, the winning attribute may be picked with a degree of confidence, as is described below. 

\begin{table}
	\caption{Hoeffding Tree, Domingos \& Hulten (2000)}
	\includegraphics[width=78.0mm]{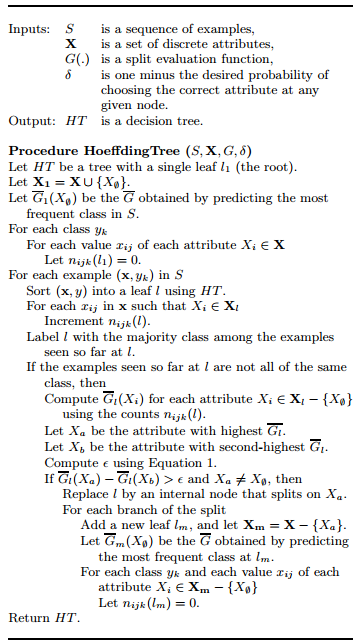} 
	\label{table:vfdt}
\end{table}

\textbf{Hoeffding Bound}: If we have $n$ independent random variables $r_1..r_n$, with range $R$ and mean $\bar{r}$, the Hoeffding bound states that with probability $1-\delta$ the true mean is at least $\bar{r} - \epsilon$ where \cite{domingos2000mining, hoeffding1963probability}:

\begin{equation} \label{eq:1}
\epsilon = \sqrt{\frac{R^2 \ln(1/\delta)}{2n}}
\end{equation} 

\textit{Hoeffding Tree} is a tree that uses this probabilistic guarantee to test at each leaf whether the computed difference of information gains $\Delta\overline{G}$ between the attributes $X_a$ and $X_b$ with highest information gains respectively, $\Delta\overline{G}(X_a)$ $-$ $\Delta\overline{G}(X_b)$, is positive and non-zero. If, for the specified tolerance $\delta$, we have $\Delta\overline{G} > \epsilon$, then we assert with confidence that $X_a$ is the better split.

Note that we are seeking to determine the best split out-of-sample. The above controls the risk that $X_a$ is inferior to $X_b$, but it does not control the risk that $X_a$ is inferior to some other attribute $X_c$. It is increasingly likely that some other split will turn out to be superior as the total number of attributes increases. There is no recourse to alter the tree in such a scenario. 

\section{Hoefdding AnyTime Tree}

If the objective is to build an incremental learner with good predictive power at any given point in the instance stream, it may be desirable to exploit information as it becomes available, building structure that improves on the current state but making subsequent corrections when further alternatives are found to be even better. In scenarios where information distribution among attributes is skewed, with some attributes containing more information than others, such a policy can be highly effective because of the limited cost of rebuilding the tree when replacing a higher-level attribute with a highly informative one. However, where information is more uniformly distributed among attributes, Hoeffding Tree will struggle to split and might have to resort to using a tie-breaking threshold that depends on the number of random variables, while HATT will pick an attribute to begin with and switch when necessary, leading to faster learning.

In this paper, we describe HATT, and provide an instantiation that we denote Extremely Fast Decision Tree (EFDT).

Hoeffding Anytime Tree is equivalent to Hoeffding tree except that it uses the Hoeffding bound to determine whether the merit of splitting on the best attribute exceeds the merit of not having a split, or the merit of the current split attribute. In practice, if no split attribute exists at a node, rather than splitting only when the top candidate split attribute outperforms the second-best candidate, HATT will split when the information gain due to the top candidate split is non-zero with the required level of confidence. At later stages, HATT will split when the difference in information gain between the current top attribute and the current split attribute is non-zero, assuming this is better than having no split. HATT is presented in Algorithm \ref{table:HATT}, Function \ref{table:attempttosplit}, and Function \ref{table:reevalsplit}.

\begin{algorithm}
	\DontPrintSemicolon
	\SetAlgoLined
	\KwIn{$S$, a sequence of examples. At time t, the observed sequence is $S^t = ((\vec{x}_1, y_1), (\vec{x}_2, y_2), ... (\vec{x}_t, y_t))$ 
		\\ \Indp\Indp $\mathbf{X} = \{X_1, X_2... X_m\}$, a set of $m$ attributes
		\\  $\delta$, the acceptable probability of choosing the wrong split attribute at a given node
		\\ G(.), a split evaluation function
	}
	\KwResult{$HATT^t$, the model at time $t$ constructed from having observed sequence $S^t$.}
	\Begin{
		Let HATT be a tree with a single leaf, the $root$ \;
		Let $\mathbf{X_1} = \mathbf{X} \cup {X_\emptyset}$\;
		Let $G_1(X_{\emptyset})$ be the $G$ obtained by predicting the most
		frequent class in S\;
		\ForEach {class $y_k$}{
			\ForEach { value $x_{ij}$ of each attribute $X_i \in \mathbf{X}$}{
				Set counter $n_{ijk}(root) = 0$\;
			}
		}
		
		\ForEach {example $(\vec{x},y)$ in S}{
			Sort $(\vec{x},y)$ into a leaf $l$ using $HATT$\;
			\ForEach {node in path $(root ... l)$}{
				\ForEach {$x_{ij}$ in $\vec{x}$ such that $X_i \in X_{node}$}{
					Increment $n_{ijk}(node)$\;
					\eIf{$node = l$}{
						$AttemptToSplit(l)$
					}
					{
						$ReEvaluateBestSplit(node)$
					}
					
				}
			}
		}
	}
	\caption{Hoeffding Anytime Tree \label{table:HATT}}
\end{algorithm}

\begin{function}
	\DontPrintSemicolon
	\SetAlgoLined
	\Begin{
		Label $l$ with the majority class at $l$\;
		\If {all examples at $l$ are not of the same class}{
			Compute $\overline{G_l}(X_i)$ for each attribute $\mathbf{X}_l - \{\mathbf{X_{\emptyset}}\}$ using the counts $n_{ijk}(l)$\;
			Let $X_a$ be the attribute with the highest $\overline{G_l}$\;
			Let $X_b = X_{\emptyset}$\;	
			Compute $\epsilon$ using equation \ref{eq:1}\;
			\If {$\overline{G_l}(X_a) - \overline{G_l}(X_b) > \epsilon$ and $X_a \neq X_{\emptyset}$}{
				Replace $l$ by an internal node that splits on $X_a$\;
				\For {each branch of the split}{
					Add a new leaf $l_m$ and let $\mathbf{X}_m = \mathbf{X} - X_a$\;
					Let $\overline{G_m}(X_{\emptyset})$ be the $G$ obtained by predicting the most frequent class at $l_m$\;
					\For {each class $y_k$ and each value $x_{ij}$ of each attribute $X_i \in X_m -$  $\{X_{\emptyset}\}$}{
						Let $n_{ijk}(l_m) = 0$.
					} 
				}		
			}
		}
	}
	\caption{AttemptToSplit(leafNode $l$)\label{table:attempttosplit}}
\end{function}

\begin{function}
	\DontPrintSemicolon
	\SetAlgoLined
	\Begin{
		Compute $\overline{G}_{int}(X_i)$ for each attribute $\mathbf{X}_{int} - \{\mathbf{X_{\emptyset}}\}$ using the counts $n_{ijk}({int})$\;
		Let $X_a$ be the attribute with the highest $\overline{G}_{int}$\;
		Let $X_{current}$ be the \textit{current} split attribute\;	
		Compute $\epsilon$ using equation \ref{eq:1}\;
		\If {$\overline{G}_l(X_a) - \overline{G}_l(X_{current}) > \epsilon$}{
			\uIf{$X_a = X_{\emptyset}$}{
				Replace internal node $int$ with a leaf (kills subtree)\;
			}
			\ElseIf{$X_a \neq X_{current}$}{
				Replace $int$ with an internal node that splits on $X_a$\;
				\For {each branch of the split}{
					Add a new leaf $l_m$ and let $\mathbf{X}_m = \mathbf{X} - X_a$\;
					Let $\overline{G}_m(X_{\emptyset})$ be the $G$ obtained by predicting the most frequent class at $l_m$\;
					\For {each class $y_k$ and each value $x_{ij}$ of each attribute $X_i \in X_m -$  $\{X_{\emptyset}\}$}{
						Let $n_{ijk}(l_m) = 0$.
					}
				}	
			}
		}

	}	
	\caption{ReEvaluateBestSplit(internalNode $int$)\label{table:reevalsplit}}
\end{function}

\subsection{Convergence}
Hoeffding Tree offers guarantees on the expected disagreement from a batch tree trained on an infinite dataset (which is denoted $DT_*$ in \cite{domingos2000mining}, a convention we will follow). ``Extensional disagreement'' is defined as the probability that a pair of decision trees will produce different predictions for an example, and intensional disagreement that probability that the path of an example will differ on the two trees.

The guarantees state that either form of disagreement is bound by $\frac{\delta}{p}$, where $\delta$ is a tolerance level and $p$ is the leaf probability-- the probability that an example will fall into a leaf at a given level. $p$ is assumed to be constant across all levels for simplicity.

Note that the guarantees will weaken significantly as the depth of the tree increases. While the built trees may have good prequential accuracy in practice on many test data streams, increasing the complexity and size of data streams such that a larger tree is required increases the chance that a wrong split is picked.

On the other hand, HATT converges in probability to the batch decision tree; we prove this below.

For our proofs, we will make the following assumption:

\begin{itemize}

	\item No two attributes will have identical information gain. This is a simplifying assumption to ensure that we can always split given enough examples, because $\epsilon$ is monotonically decreasing.

\end{itemize}

\begin{lemma}\label{lem:reaches_same_root_split_as_HT}
	
	HATT will have the same split attribute at the root as HT at the time HT splits the root node.
	
\end{lemma}	
	
\begin{proof}
	
	Let $S$ represent an infinite sequence drawn from a probability space $(\Omega, \mathcal{F}, P)$, where $(\vec{x},y) \in \Omega$ constitute our data points. The components of $\vec{x}$ take values corresponding to attributes $X_1, X_2,$ ... $X_m$, if we have $m$ attributes.
	
	We are interested in attaining confidence $1-\delta$ that $\sum_{i=0}^{n}{\Delta G}/{n} - \mu_{\Delta G} \leq \epsilon $. We don't know $\mu_{\Delta G}$, but we would like it to be non-zero, because that would imply both attributes do not have equal information gain, and that one of the attributes is the clear winner. Setting $\mu_{\Delta G}$ to $0$, we want to be confident that $\overline{\Delta G}$ differs from zero by at least $\epsilon$. In other words, we are using a corollary of Hoeffding's Inequality to state with confidence that our random variable $\overline{\Delta G}$ diverges from $0$.
	
	In order for this to happen, we need $\overline{\Delta G}$ to be greater than $\epsilon$. $\epsilon$ is monotonically decreasing, as we can see in equation \ref{eq:1}.
	
	Given the same infinite sequence of examples $S$, both HT and HATT will be presented with the same evidence $S_t(N_0)$ at the root level node $N_0$ for all $t$ (that is, indefinitely). They will always have an identical value of $\epsilon$.
	
	If at a specific time $T$ Hoeffding Tree compares attributes $X_a$ and $X_b$, which correspond to the attributes with the highest and second highest information gains $X^{1:T}$ and $X^{2:T}$ at time $T$ respectively, it follows that since $S_T(N_0)(HT) = S_T(N_0)(HATT)$, that is, since both trees have the same evidence at time $T$, Hoeffding AnyTime Tree will also find $X^{1:T} = X_a$. However, $HATT$ will compare $X_a$ with $X^T$, the current split attribute. There are four possibilities: $X^T=X^{1:T}$, $X^T=X^{2:T}$, $X^T=X^{i:T}, i > 2$ or $X^T$ is the null split. We will see that under all these scenarios, HATT will select (or retain) $X^{1:T}$.

	We need to consider the history of $\overline{\Delta G}$, which can be different for HT and HATT. That is, it is possible that for $t \leq T$, $\overline{\Delta G}(HT) \neq \overline{\Delta G}(HATT)$. This is because while HT always compares $X^{1:t}$ and $X^{2:t}$, HATT may compare $X^{1:t}$ with, say, $X^{3:t}$, $X^{4:t}$ or $X_\emptyset$, which may happen to be the current split.
	
	Clearly, at any timestep, $X^{i:t}(N_0)(HT) = X^{i:t}(N_0)(HATT)$. That is, the ranking of the information gains of the potential split attributes is always the same at the root node for both HT and HATT. It should also be obvious that since the observed sequences are identical, $G(X^{i:t}(N_0)(HT)) = G(X^{i:t}(N_0)(HATT))$-- the information gains of all of the corresponding attributes at each timestep are equal. So the top split attribute at the root $X^{1:t}(N_0)$ is always the same for both trees. If we decompose $\overline{\Delta G}^t$ as $\overline{G}_{top}^t - \overline{G}_{bot}^t$, we will have $\overline{G}_{top}^t(HT) = \overline{G}_{top}^t(HATT)$, but $\overline{G}_{bot}^t(HT)$ and $\overline{G}_{bot}^t(HATT)$ wouldn't necessarily be equal.
	
	Since at any timestep $t$ HT will always choose to compare $G(X^{1:T})$ and $G(X^{2:T})$ while HATT will always compare $G(X^{1:T})$ with $G{X_{currentSplit}}$ where $G{X_{currentSplit}} \leq G(X^{2:T})$, we have $\overline{G}_{bot}^t(HATT) \leq \overline{G}_{bot}^t(HT)$ for all $t$.

	Because we have $\overline{G}_{bot}^t(HATT) \leq \overline{G}_{bot}^t(HT)$, we will have\\ ${\overline{\Delta G}^T}(HATT) \geq {\overline{\Delta G}^T}(HT)$, and   ${\overline{\Delta G}^T}(HT) > \epsilon$ implies ${\overline{\Delta G}^T}(HATT) > \epsilon$, which would cause HATT to split on $X^{1:T}$ if it already does not happen to be the current split attribute simultaneously with HT at time $T$.
\end{proof}	

\begin{lemma} \label{lem:same_root_split_as_DT_almost_surely}
	The split attribute $X_R^{HATT}$ at the root node of HATT converges in probability to the split attribute $X_R^{DT_*}$ used at the root node of $DT_*$. That is, as the number of examples grows large, the probability that HATT will have at the root a split $X_R^{HATT}$ that matches the split $X_R^{DT_*}$ at the root node of $DT_*$ goes to 1.
\end{lemma}	

\begin{proof}
	Let us denote the attributes available at the root $X_i$ and the information gain of each attribute computed at time $t$ as $G(X_i)^t$, based on the observed sequence of examples $S^t = ((\vec{x}_1,y_1), (\vec{x}_2,y_2) ... (\vec{x}_t,y_t))$.
	
	Now, we are working under the assumption that each $X_i$ has a finite, constant information gain associated with it---$DT_*$ would not converge, and thus any guarantees about $HT$'s deviation from $DT_*$ would not hold without making this assumption. Let us denote this gain $G(X_i)^{\infty}$.
	
	This in turn implies that all pairwise differences in information gain: $\Delta G^{\infty} = G(X_a)^{\infty} - G(X_b)^{\infty}$ for any two attributes $X_a$ and $X_b$ must also be finite and constant over any given infinite dataset (from which we generate a stationary stream). 
	
	As $t \rightarrow \infty$, we expect the frequencies of our data $(\vec{x},y)$ to approach their long-term frequencies given by $P$. Consequently, we expect our measured sequences of averaged pairwise differences in information gain $\overline{\Delta G}(X_{ij})^t$ to converge to their respective constant values on the infinite dataset $\Delta G(X_{ij})^{\infty}$, which implies we will effectively have the chosen split attribute for $HATT$ converging in probability to the chosen split attribute for $DT_*$ as $t \rightarrow \infty$.
	
	Why would this convergence only be in probability and not almost surely?
	
	For any finite sequence of examples $S^t = ((\vec{x}_1,y_1),$ $(\vec{x}_2,y_2)$ $...$ $(\vec{x}_t,y_t))$ with frequencies of data that approach those given by $P$, we may observe with nonzero probability a followup sequence $((\vec{x}_{t+1}, y_{t+1}), (\vec{x}_{t+2}, y_{t+2}), ... (\vec{x}_{2t}, y_{2t}))$ that will result in a distribution that is unlike $P$ over the observations. Obviously, we expect the probability of observing such an anomalous sequence to go to $0$ as $t$ grows large-- if we didn't, we would not expect the observed frequencies of the instances to ever converge to their long-term frequencies.
	
	Anytime we do observe such a sequence, we can expect to see anomalous values of $\overline{\Delta G}(X_{ij})^t$, which means that even if the top attribute has already been established as one that matches the attribute corresponding to $\Delta G(X_{ij})^{\infty}$, it may briefly be replaced by an attribute that is not the top attribute as per $G(X_i)^{\infty}$. We have already reasoned that the probability of observing such anomalous sequences must go to $0$; so we expect that the probability of observing sequences with instance frequencies approaching those given by the measure $P$ must go to $1$. And for a sequence that is distributed as per $P$, we expect our information gain differences $\overline{\Delta G}(X_{ij})^t \rightarrow \Delta G(X_{ij})^{\infty}$.
	
	Remember that we have assumed that the pairwise differences in information gain  $\Delta G(X_{ij})^t$ are nonzero (by implication of no two attributes having identical information gain). Since $\epsilon$ is monotonically decreasing and no two attributes have been assumed to be identical, as $t$ grows large, we will always pick the attribute with the largest information gain because its advantage over the next best attribute will exceed some fixed $\epsilon$; and this picked top attribute will match, in probability, the one established by $DT_*$.

\end{proof}

\begin{lemma}\label{thm:converges_to_dt}
	Hoeffding AnyTime Tree converges to the asymptotic batch tree in probability.
	
\end{lemma}	
	
\begin{proof}
	From Lemma \ref{lem:same_root_split_as_DT_almost_surely}, we have that as $t \rightarrow \infty$, $X_R^{HATT} \xrightarrow{P} X_R^{DT_*}$, meaning that though it is possible to see at any individual timestep $X_R^{HATT} \neq X_R^{DT_*}$, we have have convergence in probability in the limit.	
	
	Consider immediate subtrees of the root node $HATT^1_i$ (denoting they are rooted at level 1). In all cases where the root split matches $X_R^{DT_*}$, the instances observed at the roots of $HATT^1_i$ will be drawn from the same data distribution that the respective $DT_{*i}^1$ draw their instances from. Do level 1 split attributes for HATT,  $X_{i:L1}^{HATT}$ converge to $X_{i:L1}^{DT_*}$?
	
	We can answer this by using the Law of Total Probability. Let us denote the event that for  first level split $i$,  $X_{i:L1}^{HATT} = X_{i:L1}^{DT_*}$ by $match_{i:L1}$. Then we have as $t \rightarrow \infty$:

	\begin{gather*}
P(X_{i:L1}^{HATT} = X_{i:L1}^{DT_*}) \\
= P(match_{i:L1}) \\
=P(match_{i:L1}|match_{L0}) P(match_{L0}) \\
	+ P(match_{i:L1}|{not\_match_{L0}}) P(not\_match_{L0})
	\end{gather*}
	
	We know that $P(match_{L0}) \rightarrow 1$ and $P(not\_match_{L0}) \rightarrow 0$ as $t \rightarrow \infty$ from Lemma \ref{lem:reaches_same_root_split_as_HT}. So we obtain
	$P(X_{i:L1}^{HATT} = X_{i:L1}^{DT_*})^\infty = P(match_{i:L1}|match_{L0})^\infty$.
	
	Effectively, we end up only having to  condition on the event $match_{L0}$. In other words, we may safely use a subset of the stream where only $match_{L0}$ has occurred to reason about whether $X_{i:L1}^{HATT} = X_{i:L1}^{DT_*}$ as $t \rightarrow \infty$.

	Now, we need to show that $P(match_{i:L1}|match_{L0}) \rightarrow 1$ as $t \rightarrow \infty$ to prove convergence at level 1. This is straightforward. Since we are only considering instances that result in the event $match_{L0}$ occurring, the conditional distributions at level 1 of HATT match the ones at level 1 of $DT_*$. We may extend this argument to any number of levels; thus $HATT$ converges in probability to $DT_*$.
	

\end{proof}



\subsection{Time and Space Complexity}

\textbf{Space Complexity:} On nominal with data with $d$ attributes, $v$ values per attribute, and $c$ classes, HATT requires $O(dvc)$ memory to store node statistics at each node, as does HT \cite{domingos2000mining}. Because the number of nodes increases geometrically, there may be a maximum of $(1-v^d)/(1-v)$ nodes, and so the worst case space complexity is O($v^{d-1}dvc$). Since the worst case space complexity for HT is given in terms of the current number of leaves $l$  as O($ldvc$) \cite{domingos2000mining}, we may write the space complexity for HATT as $O(ndvc)$, where $n$ is the total number of nodes. Note that $l$ is $O(n)$, so space complexity is equivalent for HATT and HT.

\textbf{Time Complexity:} There are two primary operations associated with learning for HT: (i) incorporating a training example by incrementing leaf statistics and (ii) evaluating potential splits at the leaf reached by an example. The same operations are associated with HATT, but we also increment internal node statistics and evaluate potential splits at internal nodes on the path to the relevant leaf.

At any leaf for HT and at any node for HATT, no more than $d$ attribute evaluations will have to be considered. Each attribute evaluation at a node requires the computation of $v$ information gains. Each information gain computation requires $O(c)$ arithmetic operations, so each split re-evaluation will require $O(dvc)$ arithmetic operations at each node. As for incorporating an example, each node the example passes through will require $dvc$ counts updated and thus $O(dvc)$ associated arithmetic operations. The cost for updating the node statistics for HATT is $O(hdvc)$, where $h$ is the maximum height of the tree, because up to $h$ nodes may be traversed by the example, while it is $O(dvc)$ for HT, because only one set of statistics needs to be updated. Similarly, the worst-case cost of split evaluation at each timestep is $O(dvc)$ for HT and $O(hdvc)$ for HATT, as one leaf and one path respectively have to be evaluated.


\section{Related Work}\label{sec:relatedwork}

There is a sizable literature that adapts HT in sometimes substantial ways \cite{jin2003efficient, Gama:2003:ADT:956750.956813, rutkowski2013decision} that do not, to the best of our knowledge, lead to the same fundamental change in learning premise as does HATT. \cite{rutkowski2013decision} and \cite{jin2003efficient} substitute the Hoeffding Test with McDiarmid's and the ``Normal'' test respectively; \cite{Gama:2003:ADT:956750.956813} adds support for Naive Bayes at leaves. Methods proposed prior to HT are either significantly less tight compared to HT in their approximation of a batch tree \cite{gratch1996sequential} or unsuitable for noisy streams and prohibitively computationally expensive \cite{utgoff1989incremental}.

The most related other works are techniques that seek to modify a tree through split replacement, usually for concept drift adaptation.

Drift adaptation generally requires explicit forgetting mechanisms in order to update the model so that it is relevant to the most recent data; this usually takes the form of a moving window that forgets older examples or a fading factor that decays the weight of older examples. In addition, when the underlying model is a tree, drift adaptation can involve subtree or split replacement.

Hulten et al \cite{hulten2001mining} follow up on the Hoeffding Tree work with a procedure for drift adaptation (Concept-adapting Very Fast Decision Tree, CVFDT). CVFDT has a moving window that diminishes statistics recorded at a node due to an example that has fallen out of a window at a given time step. The example statistics at each internal node change as the window moves, and existing splits are replaced if the split attribute is no longer the winning attribute and one of a set of alternate subtrees grown by splitting on winning attributes registers greater accuracy.

The idea common to both CVFDT and HATT is that of split re-evaluation. However, the circumstances, objectives, and methods are entirely different. CVFDT is explicitly designed for a drifting scenario; HATT for a stationary one. CVFDT's goal is to reduce prequential error for the current window in the expectation that this is the best way to respond to drift; HATT's goal is reduce prequential error overall for a stationary stream so that it asymptotically approaches that of a batch learner. CVFDT builds and substitutes alternate subtrees; HATT does not. CVFDT deliberately employs a range of forgetting mechanisms; HATT only forgets as a side effect of replacing splits---when a subtree is discarded, so too are all the historical distributions recorded therein. CVFDT always compares the top attributes, while HATT compares with either the current split attribute or the null split.

However, CVFDT is not incompatible with the core idea of Hoeffding Anytime Tree; it would be interesting to examine whether the idea of comparing with the null split or the current split attribute when applied to CVFDT will boost its performance on concept drifting streams. However, that is beyond the scope of this paper.

In order to avoid confusion, we will also mention the Hoeffding Adaptive Tree (HAT) \cite{bifet2009adaptive}. This method builds a tree that grows alternate subtrees if a subtree is observed to have poorer prequential accuracy on more recent examples, and substitutes an alternate when it has better accuracy than the original subtree. HAT uses an error estimator, such as ADWIN \cite{bifet2007learning} at each node to determine whether the prediction error due to a recent sequence of examples is significantly greater than the prediction error from a longer historical sequence so it can respond to drift. HATT, on the other hand, does not rely on prediction results or error, and does not aim to deliberately replace splits in response to drift. 

\section{Performance}\label{sec:performance}

Our EFDT implementation was built by changing the split evaluations of the MOA implementation of VFDT \cite{bifet2010moa}. We compared VFDT and EFDT on all UCI \cite{Lichman:2013} classification data sets with over $200,000$ instances that had an obvious classification target variable, did not require text mining, and did not contain missing values (MOA has limited support for handling missing values). To augment this limited collection of large datasets, we also studied performance on the WISDM dataset \cite{Kwapisz10activityrecognition}. In all, we have 12 benchmark datasets with a mixture of numeric and nominal attributes ranging from a few dimensions to hundreds of dimensions.

Many UCI datasets are ordered. VFDT and EFDT are both designed to converge towards the tree that would be learned by a batch learner if the examples in a stream are drawn i.i.d.{} from a stationary distribution. The ordered UCI datasets do not conform to this scenario, so we also study performance when they are shuffled in order to simulate it. To this end, we shuffled the data 10 times with the Unix \emph{shuf} utility seeded by a reproducible stream of random bytes \cite{randomGNU} to create 10 different streams, averaged our prequential accuracy results over the streams, as well as comparing with performance on the corresponding unshuffled stream.

Our experiments are easily reproducible. Instructions for processing datasets, source code for VFDT and EFDT to be used with MOA, and Python scripts to run the experiments are all available at [ \url{https://github.com/chaitanya-m/kdd2018.git}{} ].

\begin{figure}

	\subfloat[a][10 stream shuffled average.]{
		\includegraphics[height=1.18in, width=3.6in]{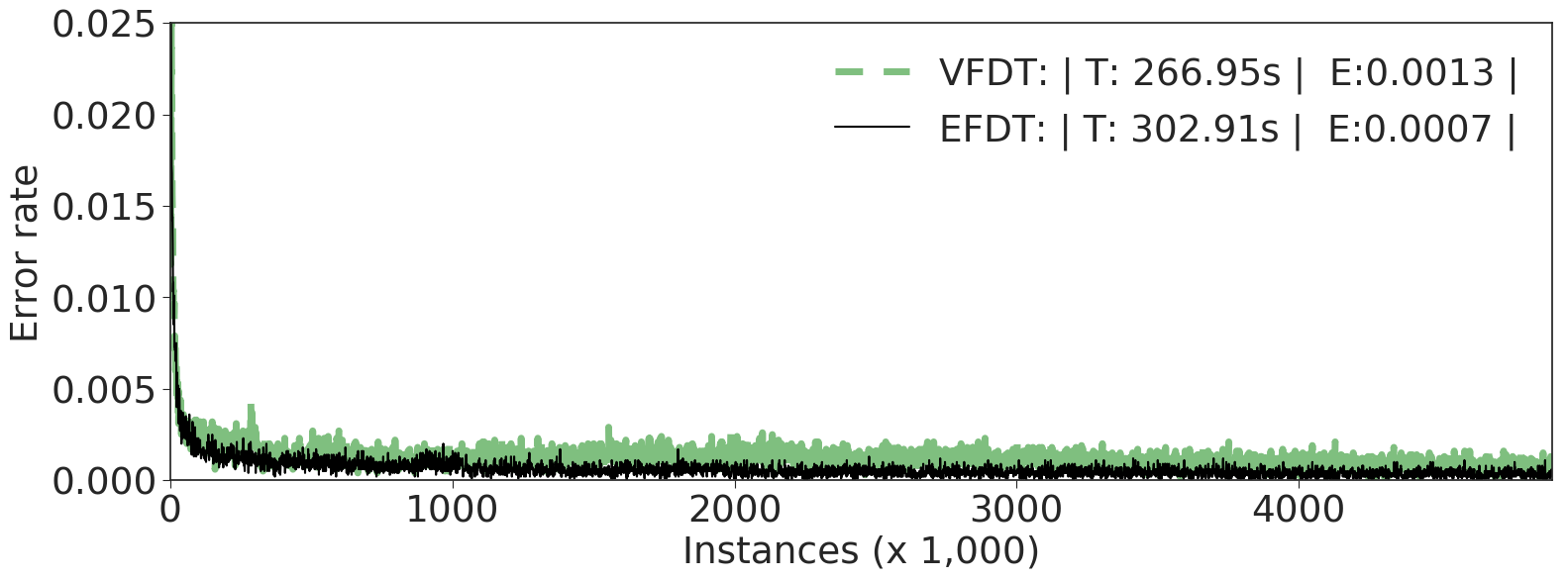} 
		}

\vspace*{-5pt}\subfloat[b][Unshuffled.]{
	\includegraphics[height=1.18in, width=3.6in]{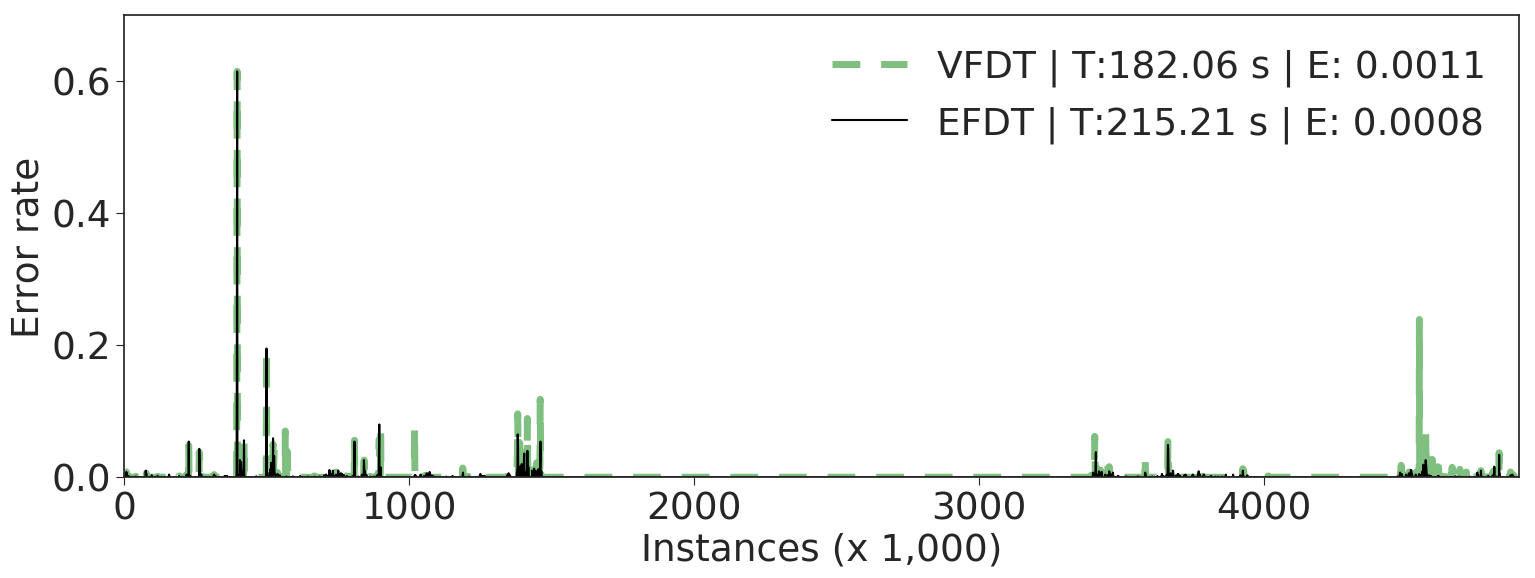} }
	\caption{KDD intrusion detection dataset \cite{Lichman:2013}}	
	\label{fig:kdd98}

\end{figure}

\begin{figure}

	\subfloat[a][ 10 stream shuffled average.]{
	\includegraphics[height=1.18in, width=3.6in]{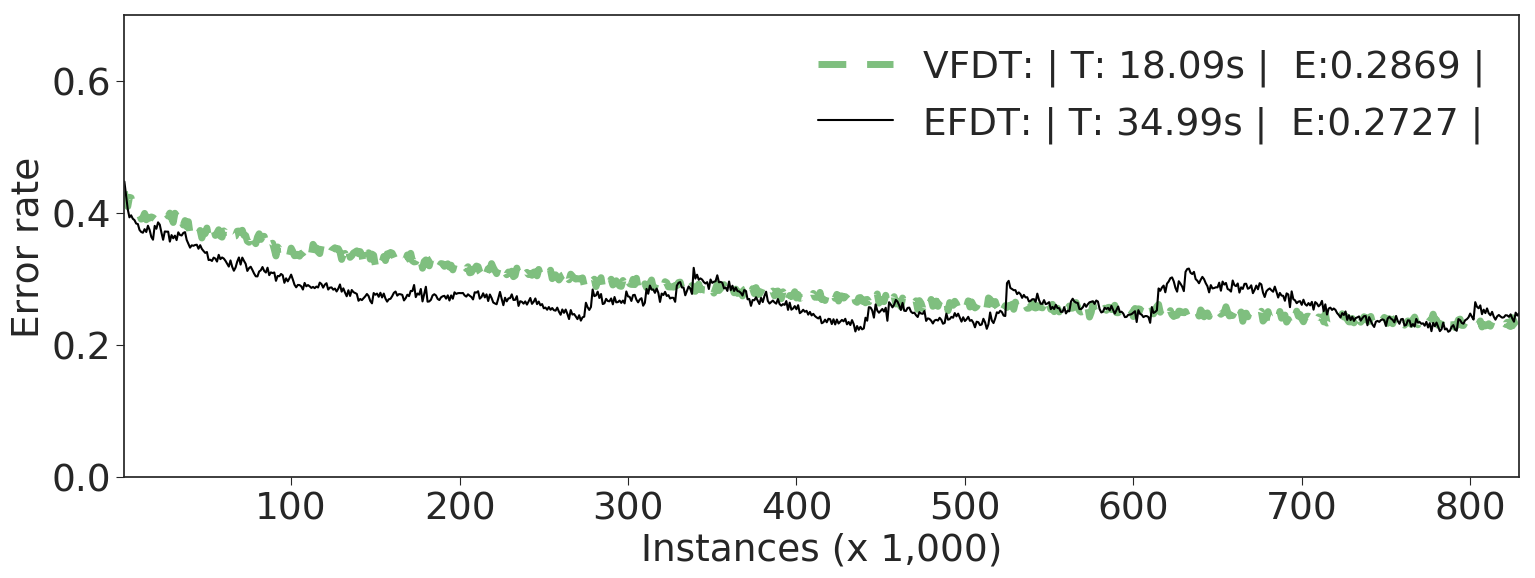} }

	\vspace*{-5pt}\subfloat[b][Unshuffled.]{\includegraphics[height=1.18in, width=3.6in]{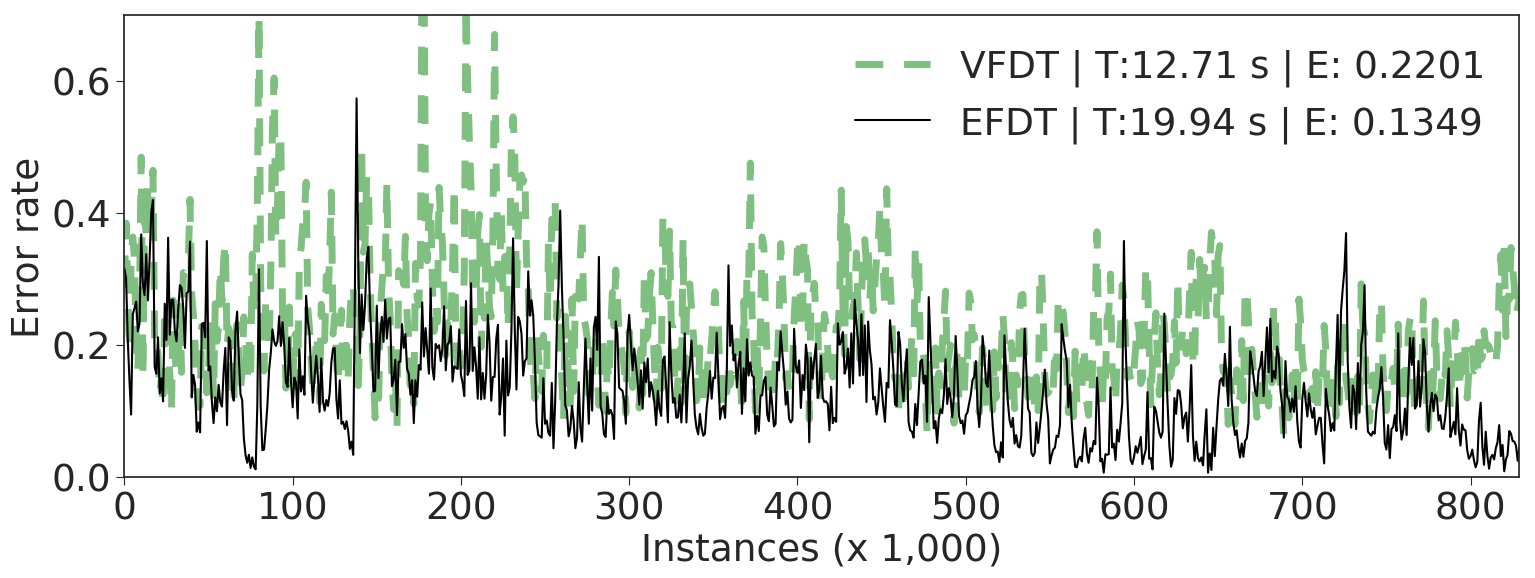} }
	\caption{Poker dataset \cite{Lichman:2013}}
	\label{fig:poker}
\end{figure}

\begin{figure}

	\subfloat[a][10 stream shuffled average.]{
	\includegraphics[height=1.18in, width=3.6in]{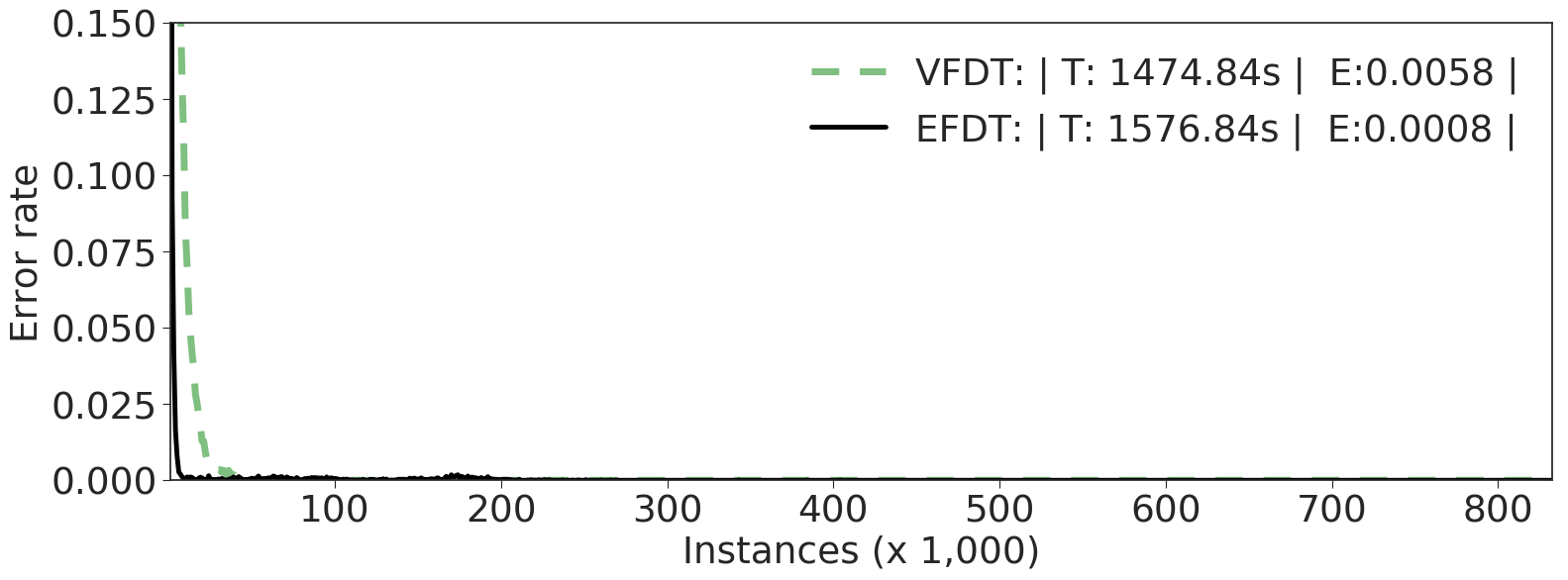} }


	\vspace*{-5pt}\subfloat[b][Unshuffled.]{
	\includegraphics[height=1.18in, width=3.6in]{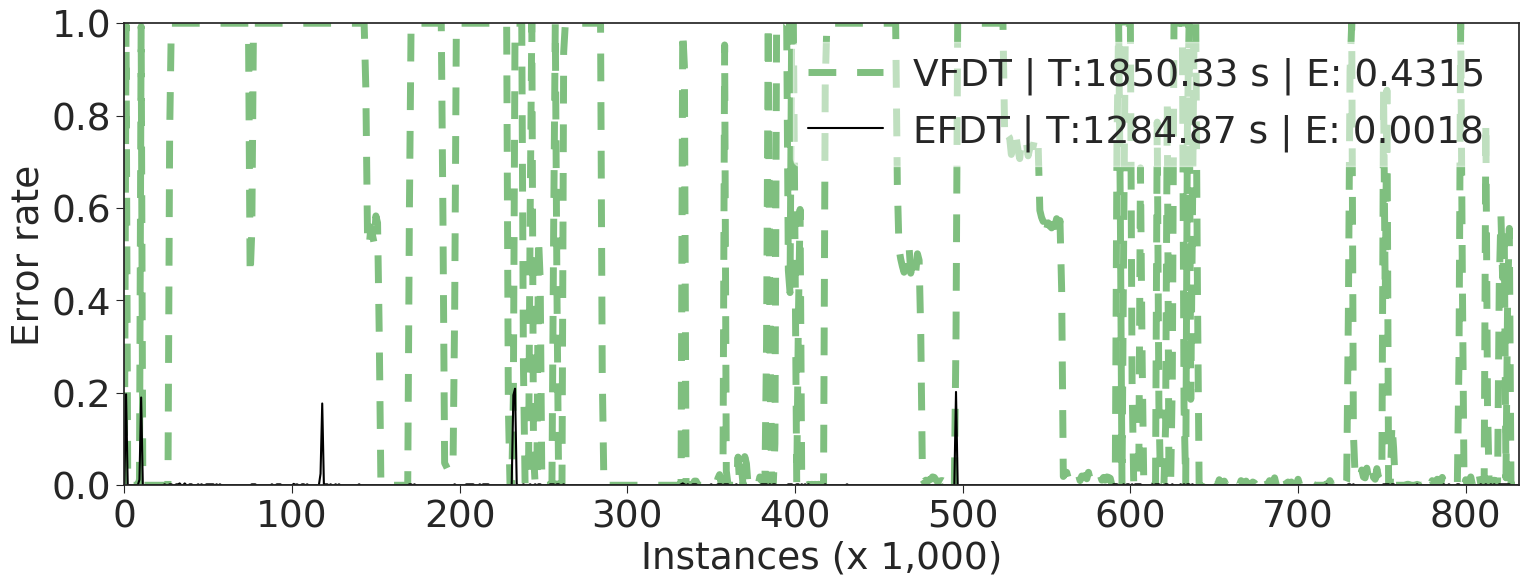} }
		\caption{Fonts dataset \cite{Lichman:2013}}
	\label{fig:font}
\end{figure}

\begin{figure}

	\subfloat[a][10 stream shuffled average.]{
	\includegraphics[height=1.18in, width=3.6in]{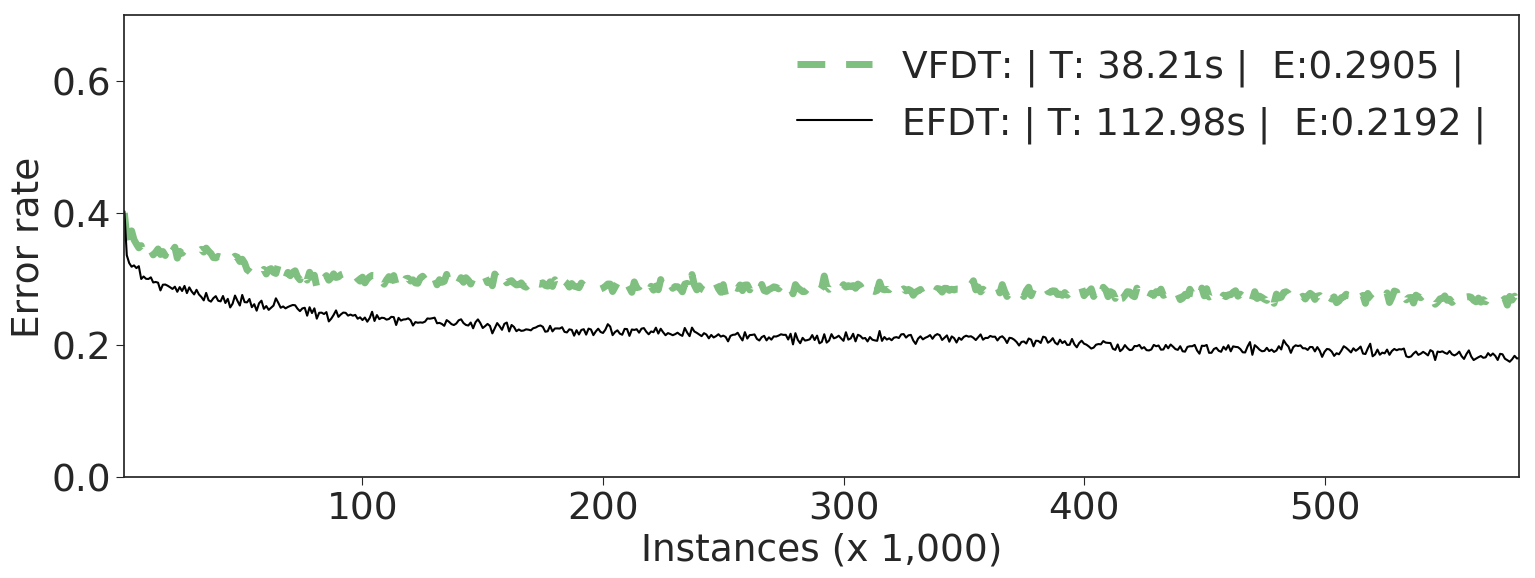} }

	\vspace*{-5pt}\subfloat[b][Unshuffled.]{
	\includegraphics[height=1.18in, width=3.6in]{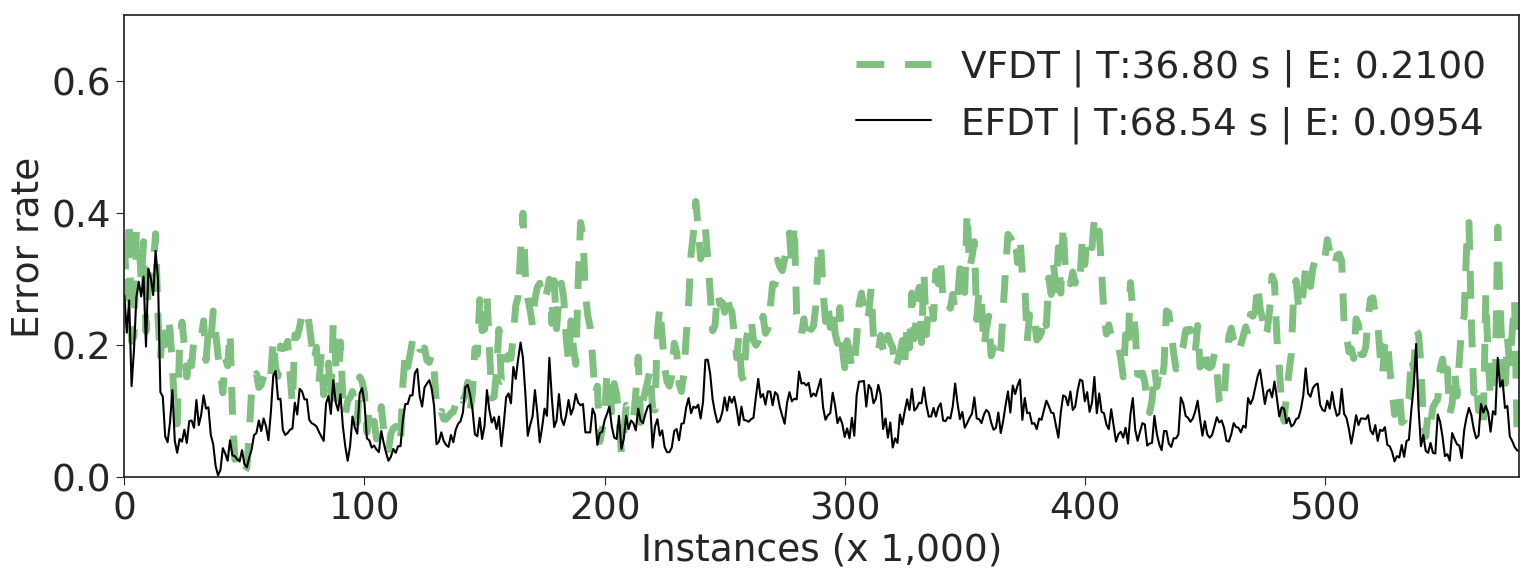} }
		\caption{Forest covertype dataset \cite{covtypedataset, Lichman:2013}}
	\label{fig:covtype}
\end{figure}

\begin{figure}

	\subfloat[a][ 10 stream shuffled average.]{
	\includegraphics[height=1.18in, width=3.6in]{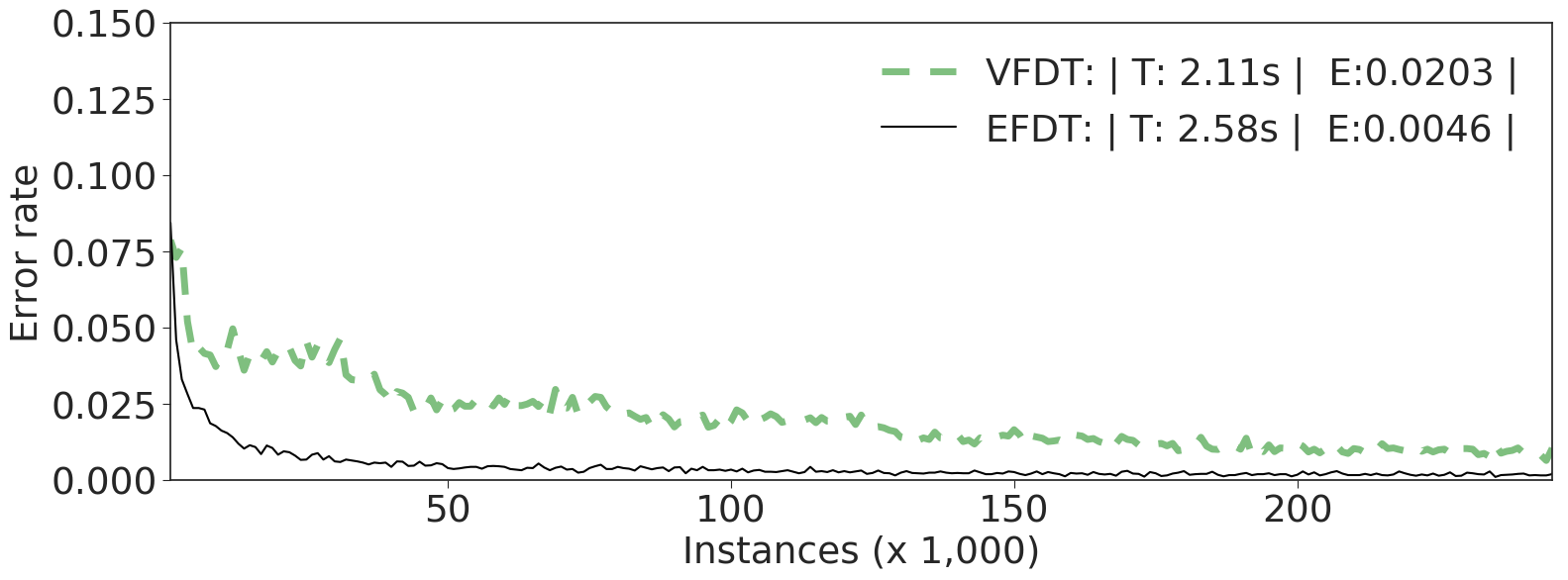} }

	\vspace*{-5pt}\subfloat[b][Unshuffled.]{
	\includegraphics[height=1.18in, width=3.6in]{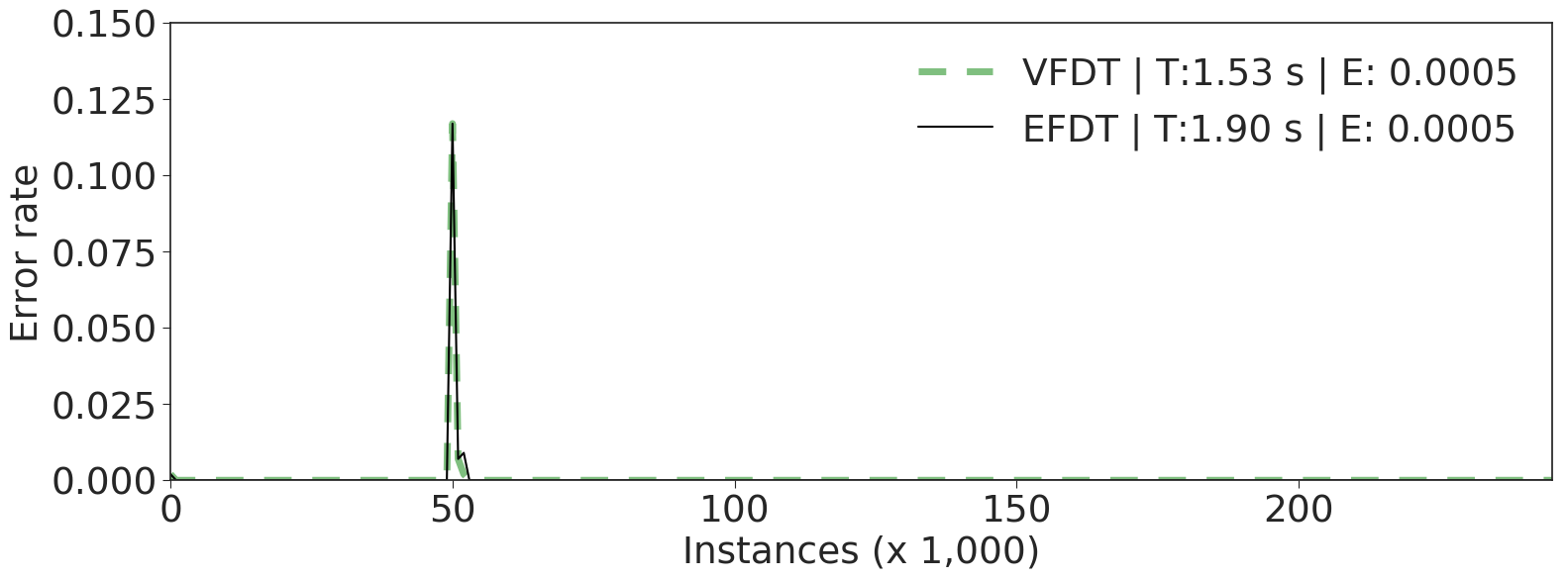} }
		\caption{Skin dataset \cite{datasetskin, Lichman:2013}}
	\label{fig:skin}
\end{figure}

\begin{figure}

	\subfloat[a][ 10 stream shuffled average.]{
		\includegraphics[height=1.18in, width=3.6in]{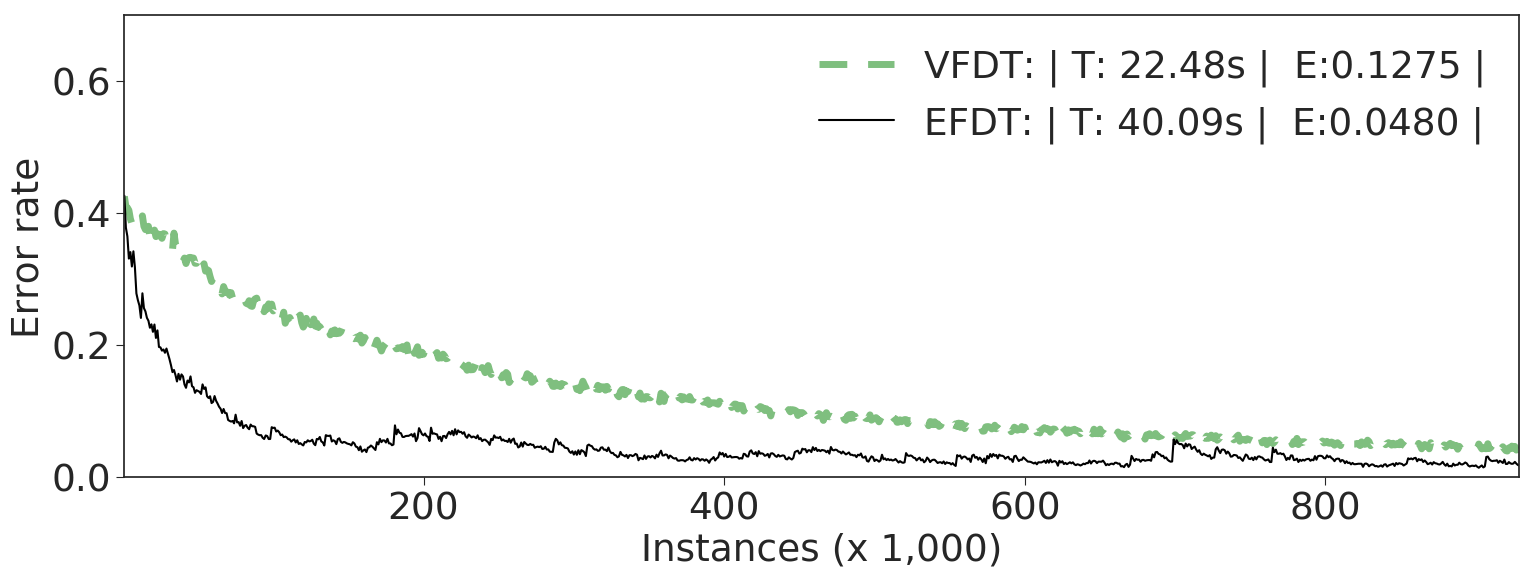} }
	
	\vspace*{-5pt}\subfloat[b][Unshuffled.]{
		\includegraphics[height=1.18in, width=3.6in]{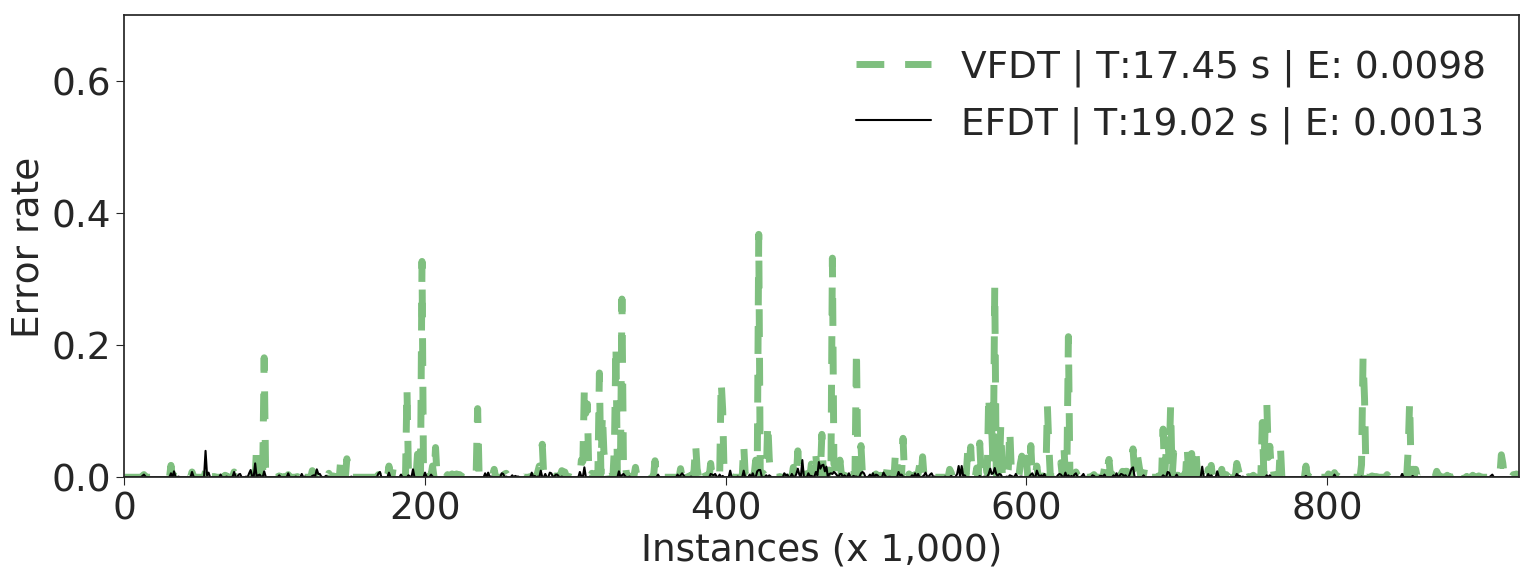} }
		\caption{Gas sensor dataset \cite{huerta2016online, Lichman:2013}}
	\label{fig:gas}
\end{figure}

\begin{figure}

		\subfloat[a][ 10 stream shuffled average.]{
	\includegraphics[height=1.18in, width=3.6in]{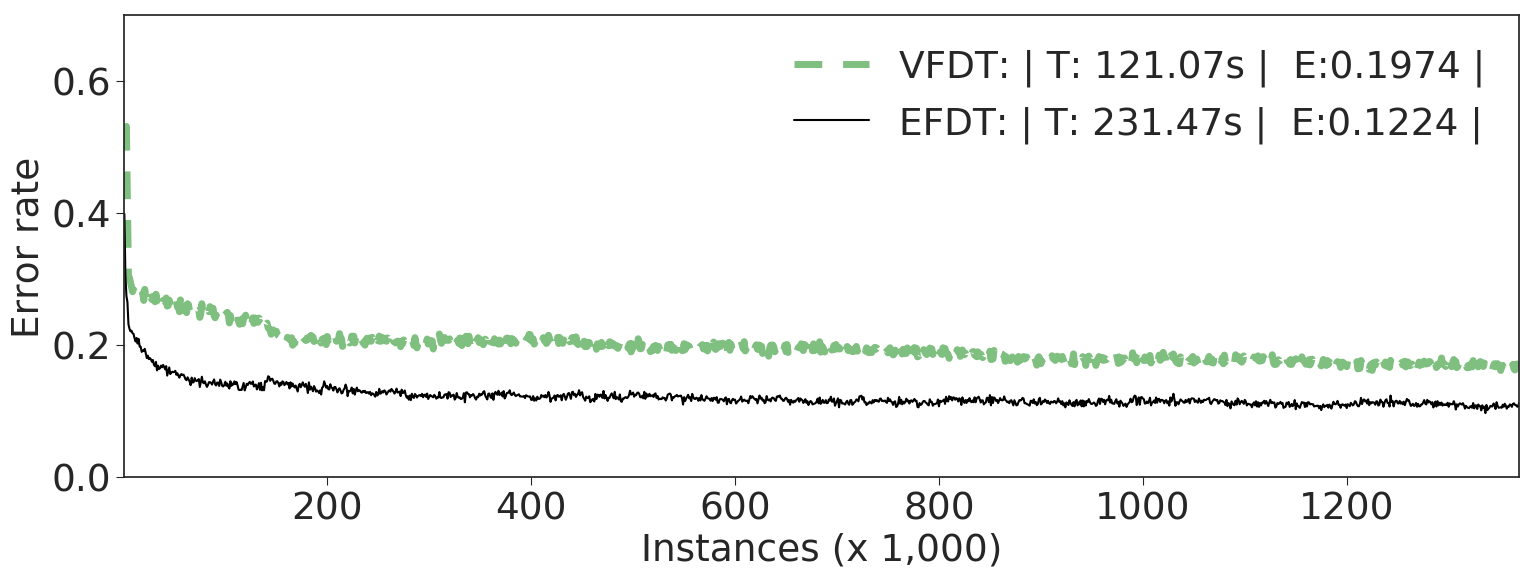} }

	\vspace*{-5pt}\subfloat[b][Unshuffled.]{
	\includegraphics[height=1.18in, width=3.6in]{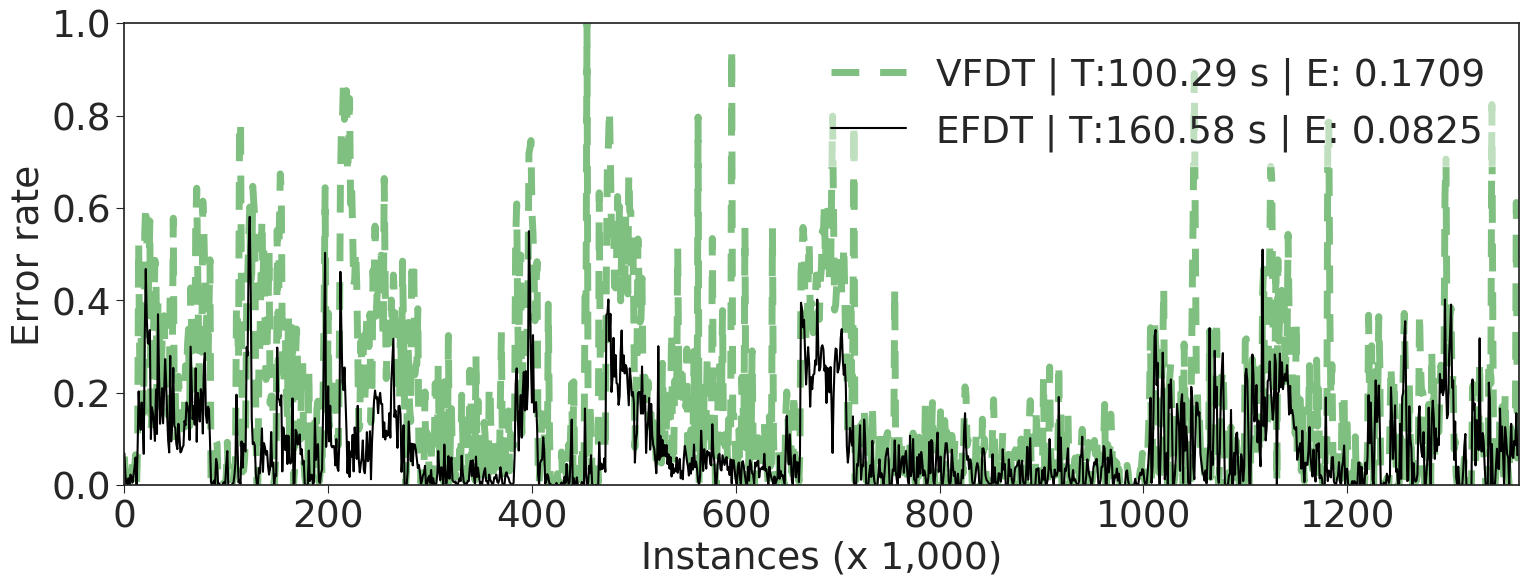} }
		\caption{WISDM dataset \cite{Kwapisz10activityrecognition, Lichman:2013}}
	\label{fig:wisdm}
\end{figure}

\begin{figure}

			\subfloat[a][ 10 stream shuffled average.]{
	\includegraphics[height=1.18in, width=3.6in]{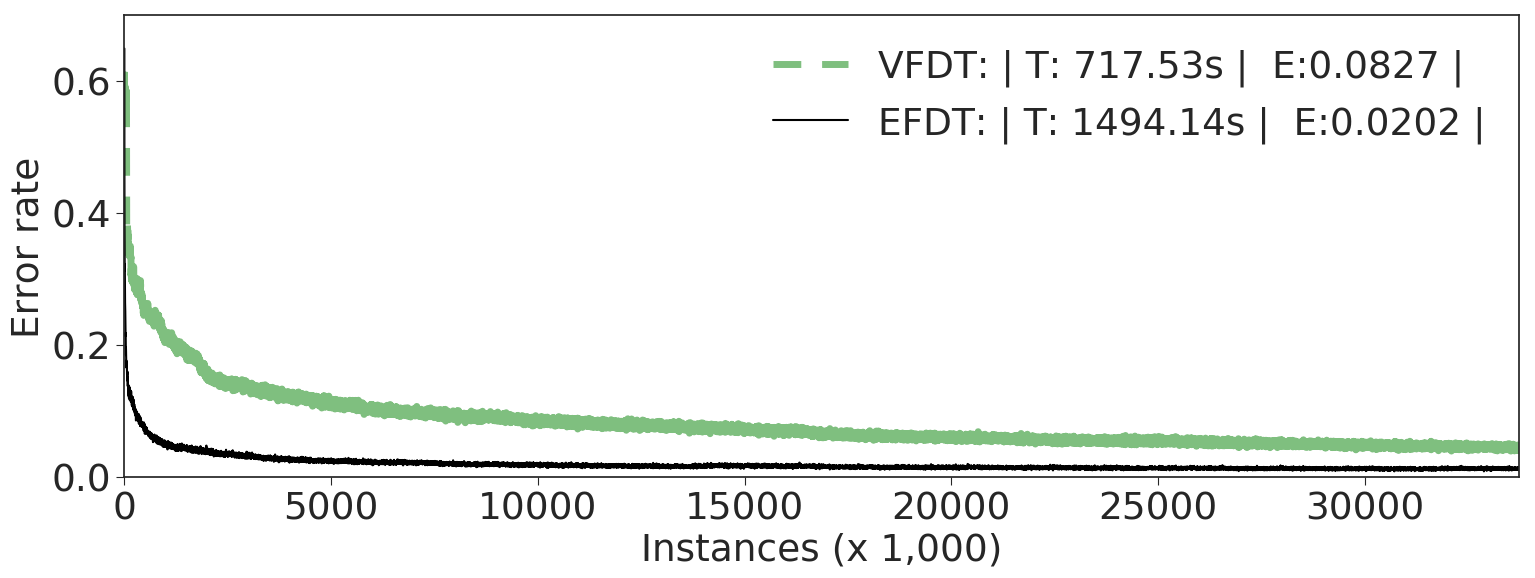} }

	\vspace*{-5pt}\subfloat[b][Unshuffled.]{
	\includegraphics[height=1.18in, width=3.6in]{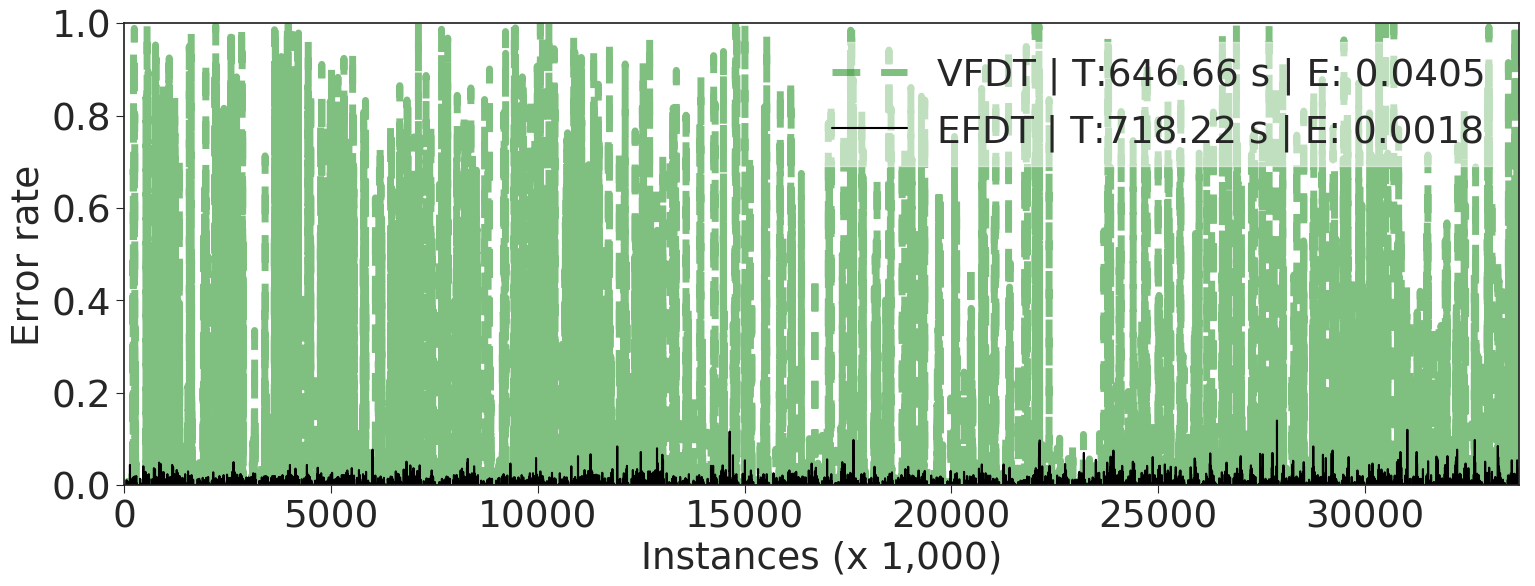} }
		\caption{Human Activity Recognition dataset: Phone, watch accelerometer, and gyrometer data combined. \cite{Stisen:2015:SDD:2809695.2809718, Lichman:2013}}
	\label{fig:har}
\end{figure}

\begin{figure}

				\subfloat[a][ 10 stream shuffled average.]{	
	\includegraphics[height=1.18in, width=3.6in]{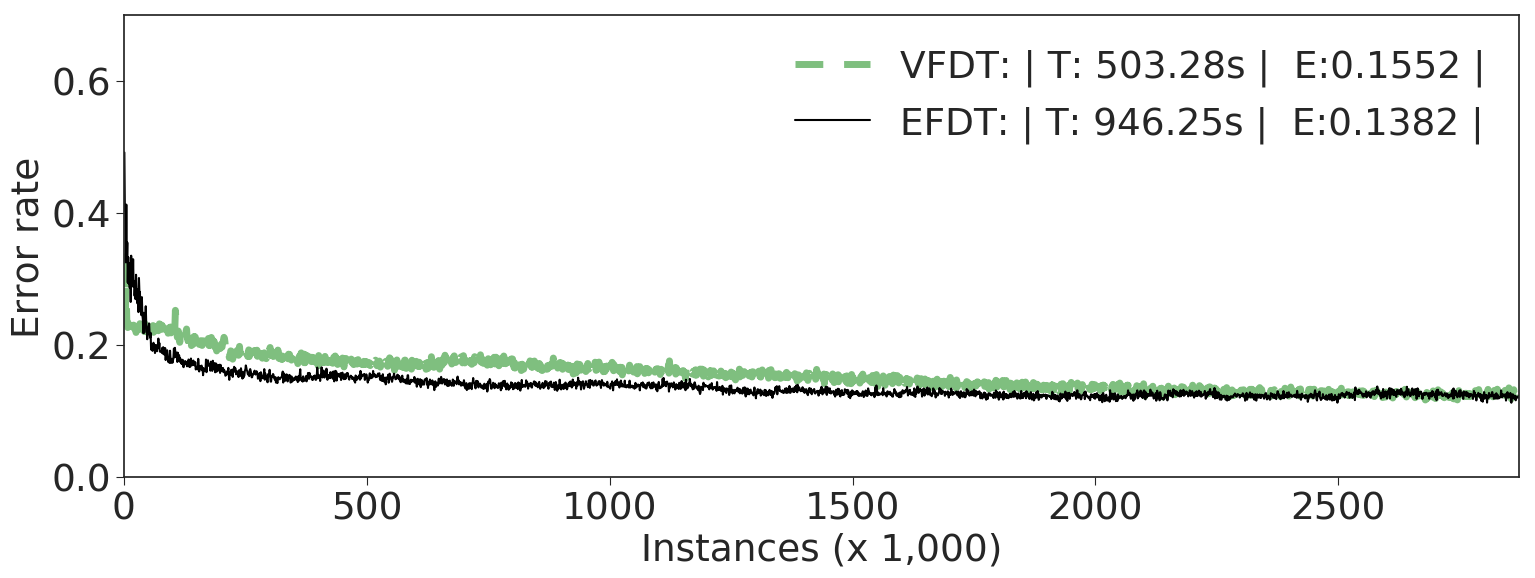} }

	\vspace*{-5pt}\subfloat[b][Unshuffled.]{
	\includegraphics[height=1.18in, width=3.6in]{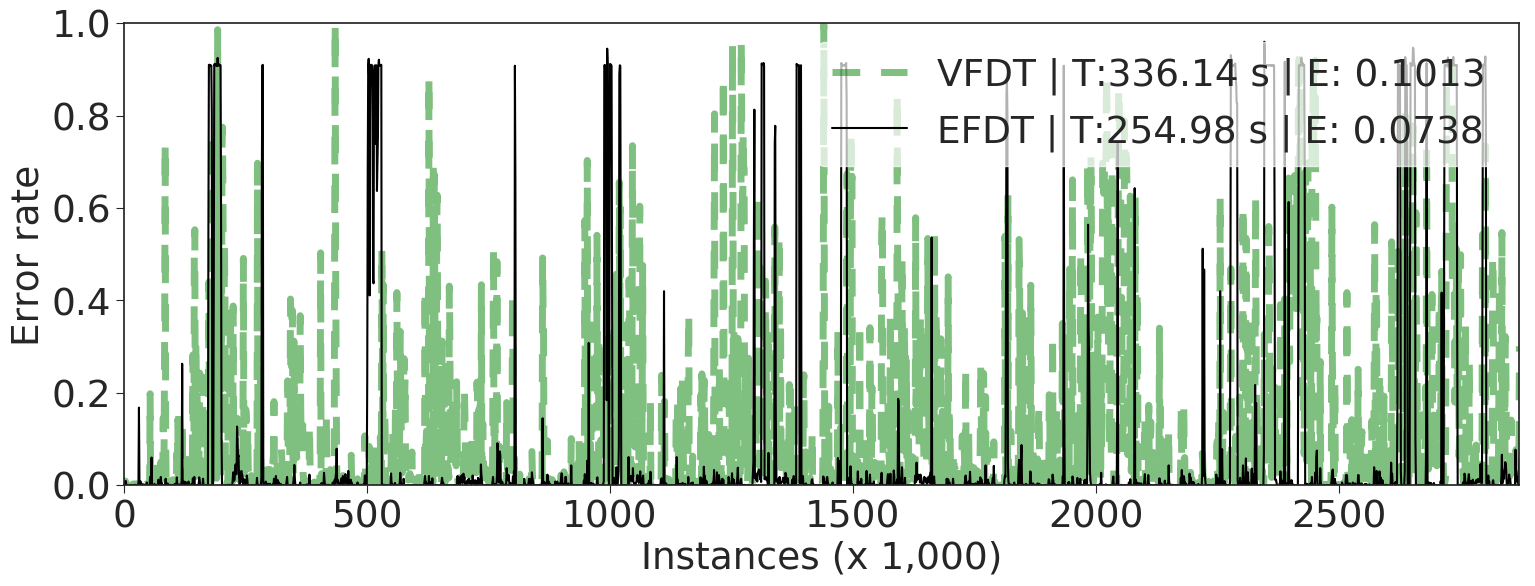} }

	\caption{PAMAP2 Activity Recognition dataset (UCI)--  9 subjects data combined \cite{reiss2012introducing, Lichman:2013}}
	\label{fig:pamap2}
\end{figure}

\begin{figure}

	\subfloat[a][ 10 stream shuffled average.]{	
	\includegraphics[height=1.18in, width=3.6in]{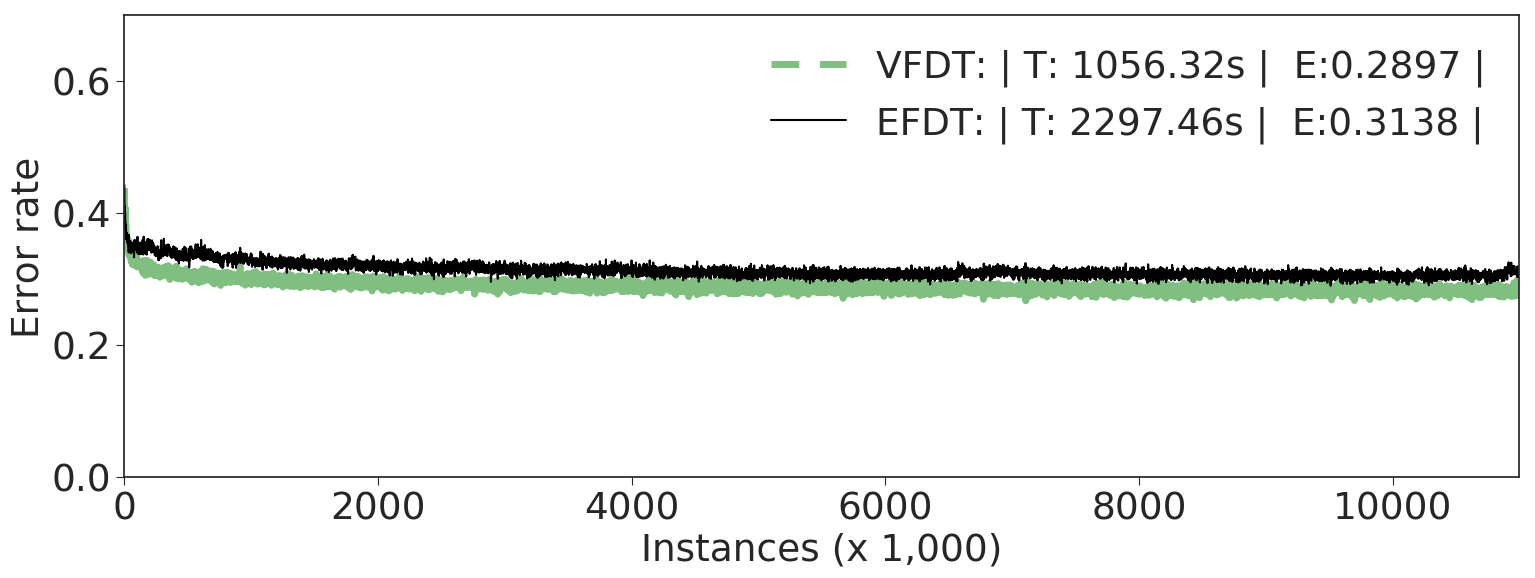} }

	\vspace*{-5pt}\subfloat[b][Unshuffled.]{
	\includegraphics[height=1.18in, width=3.6in]{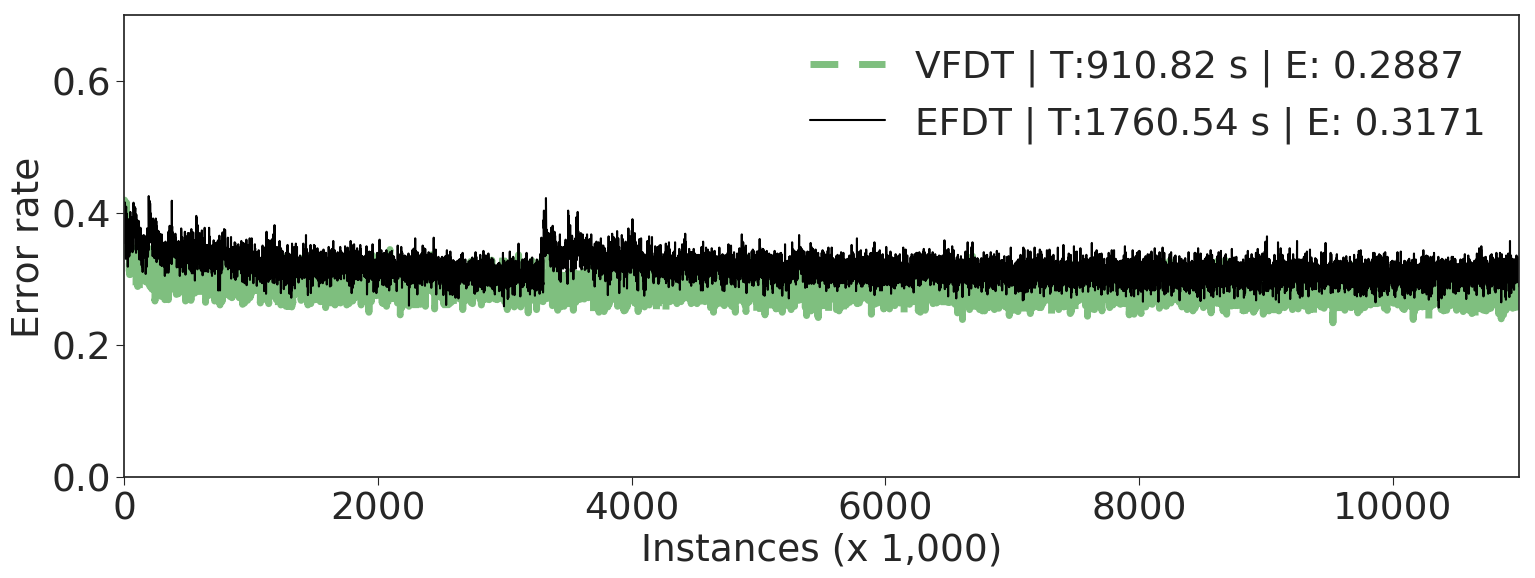} }
		\caption{Higgs dataset \cite{baldi2014searching,Lichman:2013}}
	\label{fig:higgs}
\end{figure}

\begin{figure}

	\subfloat[a][ 10 stream shuffled average.]{	
		\includegraphics[height=1.18in, width=3.6in]{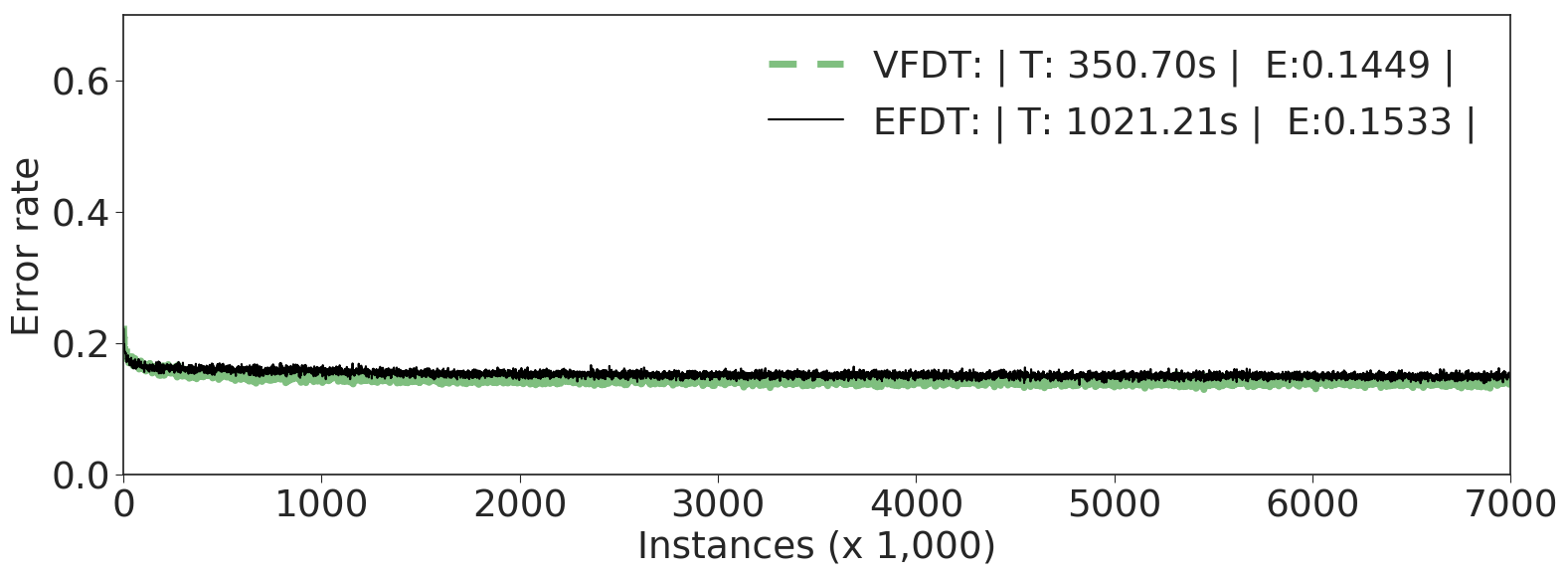} }
	
	\vspace*{-5pt}\subfloat[b][Unshuffled.]{
		\includegraphics[height=1.18in, width=3.6in]{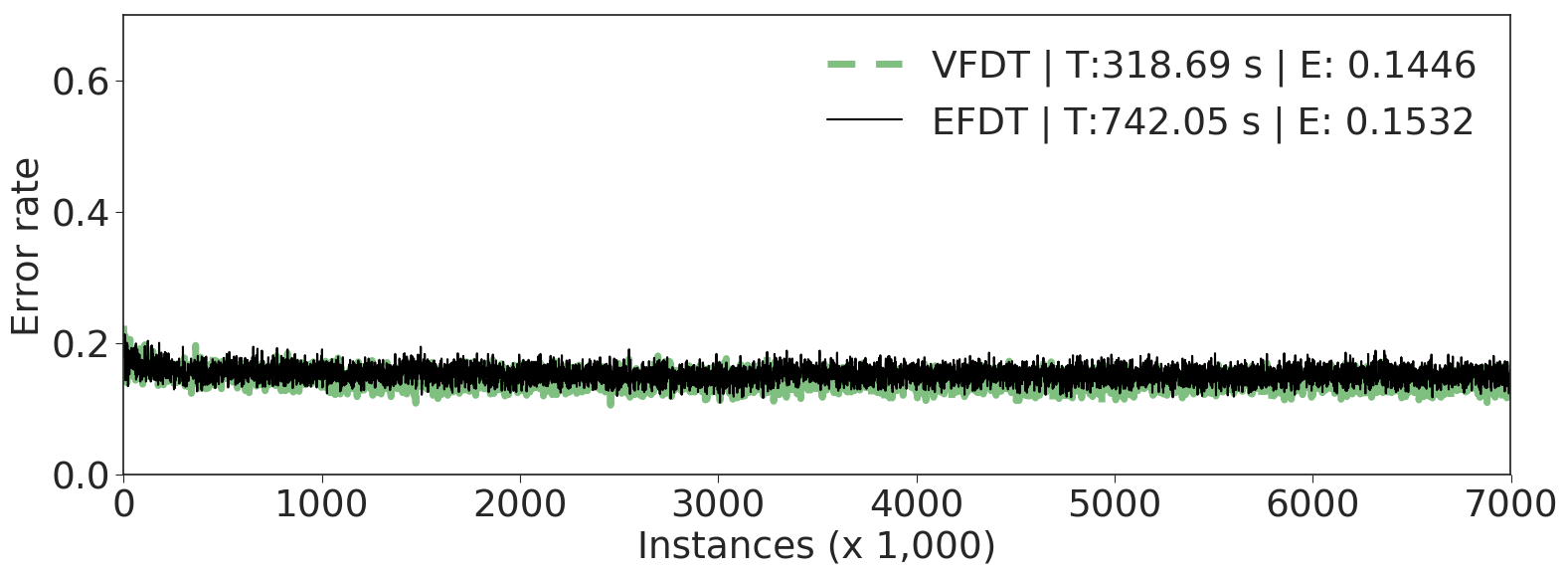} }
		\caption{Hepmass dataset \cite{baldihepmass,Lichman:2013}}
	\label{fig:hepmass}
\end{figure}

\begin{figure}

	\subfloat[a][ 10 stream shuffled average.]{	
		\includegraphics[height=1.18in, width=3.6in]{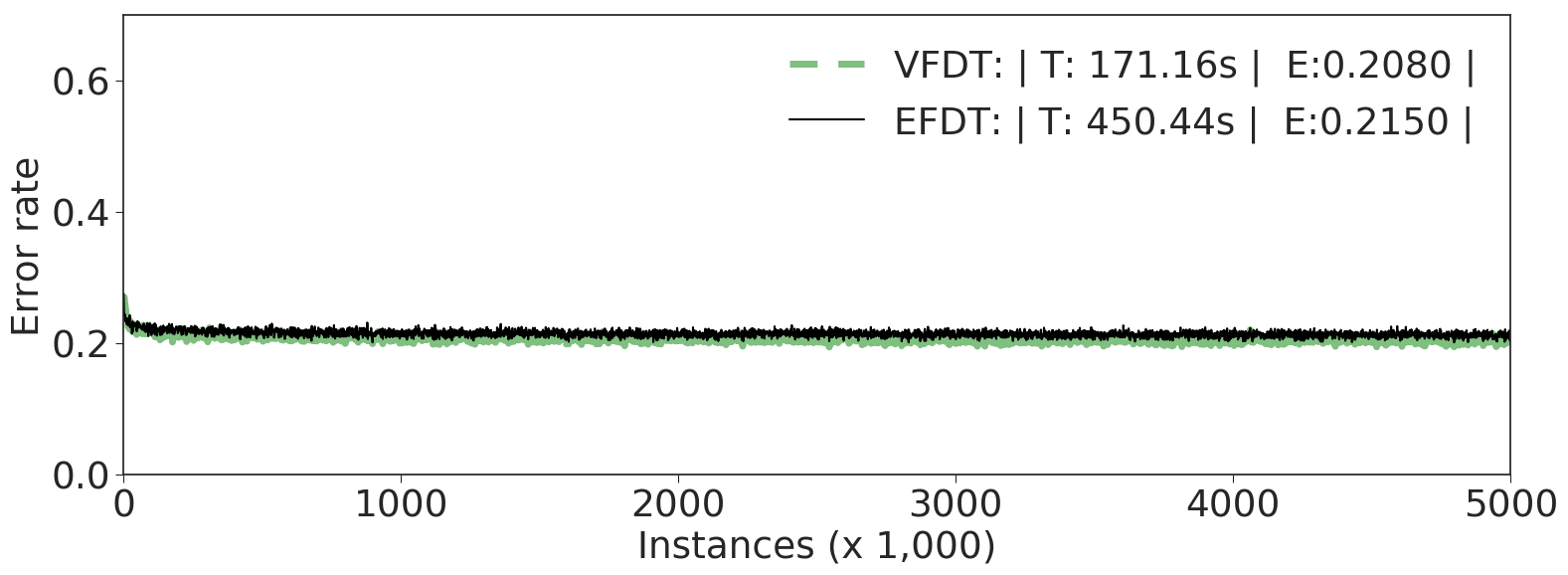} }
	
	\vspace*{-5pt}\subfloat[b][Unshuffled.]{
		\includegraphics[height=1.18in, width=3.6in]{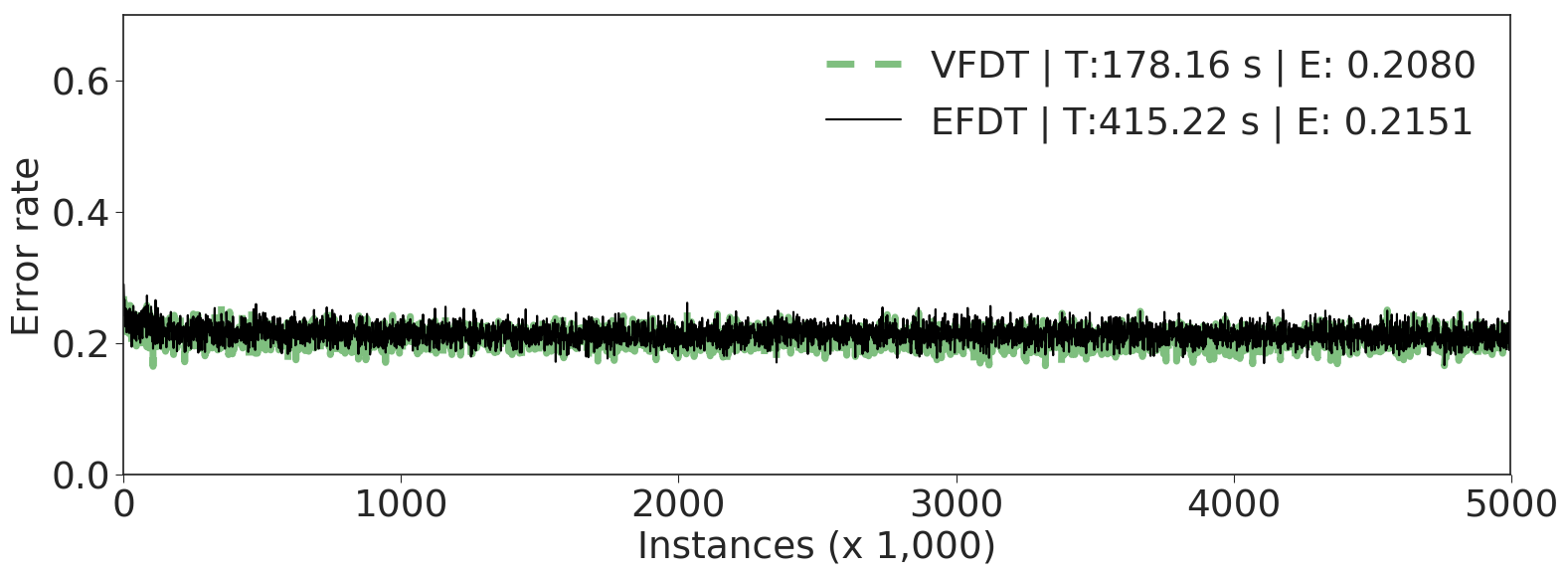} }
		\caption{SUSY dataset \cite{baldi2014searching,Lichman:2013}}
	\label{fig:susy}
\end{figure}

EFDT attains substantially higher prequential accuracy on most streams (Figs. \ref{fig:kdd98} to \ref{fig:pamap2}) whether shuffled or unshuffled. Where VFDT wins (\ref{fig:higgs}, \ref{fig:hepmass}, \ref{fig:susy}) the margin is far smaller than most of the EFDT wins. While EFDT runtime generally exceeds that of VFDT, we find it rarely requires more than double the time and in some cases, when it learns smaller trees, requires less time. We evaluate leaves every $200$ timesteps and internal nodes every $2000$ timesteps.


Differences in shuffled and unshuffled performance highlight the amount of order that is present in the unshuffled data. The unshuffled Skin dataset contains B,G,R values and a target variable that indicates whether the input corresponds to skin or not. All positive examples are at the start followed by all negative examples; the net effect is that a learner will replace one extremely simple concept with another (Fig. \ref{fig:skin}). When shuffled, it is necessary to learn a more complex decision boundary, affecting performance for both learners. 

A different effect is observed with the higher dimensional Fonts dataset (Fig. \ref{fig:font}). The goal is to predict which of 153 fonts corresponds to a 19x19 greyscale image, with each pixel able to take 255 intensity values. When instances are sorted, by font name alphabetically,each time a new font is encountered VFDT needs to learn the new concept at every leaf of an increasingly complex tree. In contrast, EFDT is able to readjust the model, efficiently discarding outdated splits to achieve an accuracy of around 99.8\%, making it a potentially powerful base learner for methods designed for concept drifting scenarios.

The results on the Poker and Forest-Covertype datasets (Figs. \ref{fig:poker}, \ref{fig:covtype}) reflect both effects: EFDT performs significantly better on ordered data, and performance for both learners deteriorates with shuffled data in comparison with unshuffled data.

Every additional level of a decision tree fragments the input space, slowing down tree growth exponentially. A delay in splitting at one level delays the start of collecting information with respect to the splits for the next level. These delays cascade, greatly delaying splitting at deeper levels of the tree.

Thus, we expect HATT to have an advantage over HT in situations where HT considerably delays splits at each level---such as when the difference in information gain between the top attributes at a node is low enough to require a large number of examples in order to overcome the Hoeffding bound, though the information gains themselves happen to be significant. This would lead to a potentially useful split in HT being delayed, and poor performance in the interim.

Conversely, when the differences in information gain between top attributes as well as the information gains themselves are low, it is possible that HATT chooses a split that would require a large number of examples to readjust. However, since we expect this to keep up with VFDT on the whole, the main source of underperformance for EFDT is likely to be an overfitted model making low-level adjustments. Synthetic data from physics simulations available in the UCI repository (Higgs, Hepmass, SUSY) led to such a scenario. 

\begin{figure}[t]
	
	\subfloat[a][VFDT]{
		\includegraphics[height=1.18in, width=3.6in]{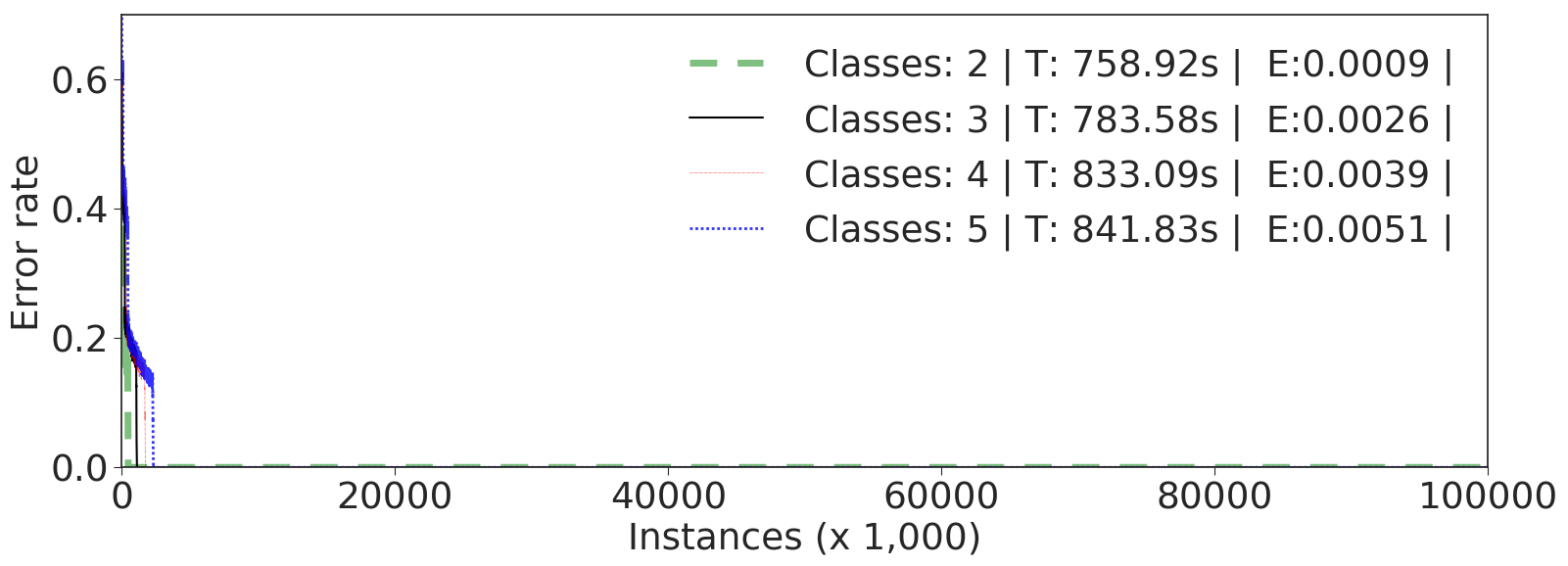}}

	\vspace*{-5pt}\subfloat[b][EFDT]{
		
		\includegraphics[height=1.18in, width=3.6in]{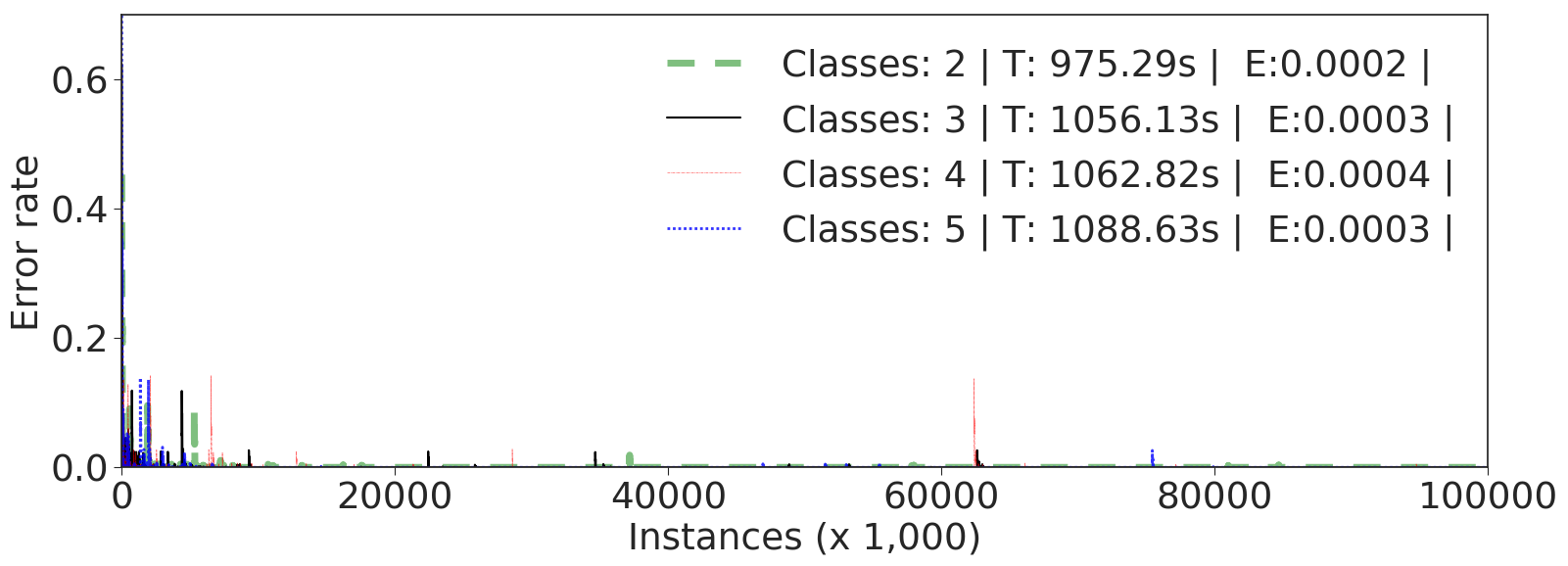} }

	\caption{A longer term view of the experiments from Fig. \ref{fig:intro} shows us that even 100 million examples in, EFDT maintains a commanding lead on prequential accuracy.}
	\label{fig:outro}
\end{figure}

Fig. \ref{fig:outro} shows us that with the MOA tree generator used in Fig. \ref{fig:intro}, even on a 100 million length stream, EFDT's prequential error is still an order of magnitude lower than that of VFDT.

\section{Conclusions}

Hoeffding AnyTime Tree makes a simple change to the current de facto standard for incremental tree learning. The current state-of-the-art Hoeffding Tree aims to only split at a node when it has identified the best possible split and then to never revisit that decision. In contrast HATT aims to split as soon as a useful split is identified, and then to replace that split as soon as a better alternative is identified. Our results demonstrate that this strategy is highly effectively on benchmark datasets.

Our experiments find that HATT has some inbuilt tolerance to concept drift, though it is not specifically designed as a learner for drift. It is easy to conceive of ensemble, forgetting, decay, or subtree replacement approaches built upon HATT to deal with concept drift, along the lines of approaches that have been proposed for HT.

HT cautiously works toward the asymptotic batch tree, ignoring, and thus not benefiting from potential improvements on the current state of the tree, until it is sufficiently confident that they will not need to be subsequently revised. If an incrementally learned tree is to be deployed to make predictions before fully learned, HATT's strategy of always utilizing the most useful splits identified to date has profound benefit.

\bibliographystyle{ACM-Reference-Format}
\bibliography{sample-bibliography} 

\end{document}